\documentclass[a4paper,onesided,12pt]{report}
\usepackage{fbe_tez}
\usepackage{xcolor}
\usepackage[utf8x]{inputenc} 

\usepackage{amsmath, amsthm, amssymb}
\usepackage[bottom]{footmisc}
\usepackage{cite}
\usepackage{url}

\usepackage{graphicx}
\usepackage{longtable}
\usepackage{fixltx2e}
\usepackage{listings}
\usepackage{algorithm}
\usepackage{algorithmic}
\usepackage{tabto}
\graphicspath{{figures/}} 

\usepackage{multirow}
\usepackage{subfigure}
\usepackage{algorithm}
\usepackage{algorithmic}

\usepackage{enumitem}

\title{Studying the impacts of pre-training using ChatGPT generated text on downstream tasks}

\turkcebaslik{Het bestuderen van de impact van pre-training met door ChatGPT gegenereerde tekst op downstream-taken}
\author{Sarthak Anand}
\program{Artificial Intelligence}
\subyear{2023}

\supervisor{Dr. Heysem Kaya}
\cosuperi{Dr. Dong Nguyen}
\dateofapproval{21.07.2023}

\begin{document}

\pagenumbering{roman}
\makemstitle 

\makeapprovalpage
\begin{acknowledgements}
\noindent I would like to express my deepest gratitude to all those who have supported and guided me throughout the journey of completing this research thesis. Firstly, I am immensely grateful to my supervisors \textbf{Dr. H. (Heysem) Kaya, G. (Gizem) Sogancioglu, Mathias Leys and Dr. D. P. (Dong) Nguyen} for their invaluable guidance, unwavering support, and constant encouragement. Their expertise, insightful feedback, and constructive criticism have been instrumental in shaping the direction and quality of this research work. I would also like to acknowledge the \textbf{ML6} for providing the necessary resources and infrastructure that facilitated the smooth progress of my research. I am indebted to my friends and colleagues who have provided constant support, encouragement, and insightful discussions throughout the research process. Specially \textbf{Rosa Eggink} for always encouraging and believing in me during the difficult phase of my thesis and \textbf{Tessel Haagen} for always motivating. Finally, I would like to express my gratitude to my family for providing me the opportunities and their belief in my abilities.  The completion of this research thesis would not have been possible without the support and contributions of all those mentioned. I am truly grateful for their presence in my journey.
\end{acknowledgements}

\begin{abstract}
In recent times, significant advancements have been witnessed in the field of language models, particularly with the emergence of Large Language Models (LLMs) that are trained on vast amounts of data extracted from internet archives. These LLMs, such as ChatGPT, have become widely accessible, allowing users to generate text for various purposes including articles, essays, jokes, and poetry. Given that LLMs are trained on a diverse range of text sources, encompassing platforms like Reddit and Twitter, it is foreseeable that future training datasets will also incorporate text generated by previous iterations of the models themselves. In light of this development, our research aims to investigate the influence of artificial text in the pre-training phase of language models. Specifically, we conducted a comparative analysis between a language model, RoBERTa, pre-trained using CNN/DailyMail news articles, and ChatGPT, which employed the same articles for its training and evaluated their performance on three downstream tasks as well as their potential gender bias, using sentiment analysis as a metric. Through a series of experiments, we demonstrate that the utilization of artificial text during pre-training does not have a significant impact on either the performance of the models in downstream tasks or their gender bias. In conclusion, our findings suggest that the inclusion of text generated by LLMs in their own pre-training process does not yield substantial effects on the subsequent performance of the models in downstream tasks or their potential gender bias.

  \end{abstract}
\begin{ozet}
De afgelopen tijd zijn er aanzienlijke vorderingen gemaakt op het gebied van taalmodellen, met name met de opkomst van Large Language Models (LLM's) die zijn getraind op enorme hoeveelheden gegevens die uit internetarchieven zijn gehaald. Deze LLM's, zoals ChatGPT, zijn algemeen toegankelijk geworden, waardoor gebruikers tekst kunnen genereren voor verschillende doeleinden, waaronder artikelen, essays, grappen en poëzie. Aangezien LLM's worden getraind op een breed scala aan tekstbronnen, waaronder platforms zoals Reddit en Twitter, is het te voorzien dat toekomstige trainingsdatasets ook tekst zullen bevatten die is gegenereerd door eerdere iteraties van de modellen zelf. In het licht van deze ontwikkeling heeft ons onderzoek tot doel de invloed van kunstmatige tekst in de pre-trainingsfase van taalmodellen te onderzoeken. We hebben met name een vergelijkende analyse uitgevoerd tussen een taalmodel, RoBERTa, vooraf getraind met behulp van nieuwsartikelen van CNN/DailyMail, en ChatGPT, dat dezelfde artikelen gebruikte voor zijn training en hun prestaties op drie stroomafwaartse taken evalueerde, evenals hun mogelijke gendervooroordelen. , met sentimentanalyse als maatstaf. Door middel van een reeks experimenten tonen we aan dat het gebruik van kunstmatige tekst tijdens de pre-training geen significante invloed heeft op de prestaties van de modellen in stroomafwaartse taken of hun gendervooroordelen. Concluderend suggereren onze bevindingen dat het opnemen van tekst gegenereerd door LLM's in hun eigen pre-trainingsproces geen substantiële effecten heeft op de daaropvolgende prestaties van de modellen in stroomafwaartse taken of hun mogelijke gendervooroordelen.
\end{ozet}
\tableofcontents
\listoffigures
\listoftables


\begin{abbreviations}
 \sym{AAE}{African American English}
 \sym{ALBERT}{A Lite BERT}
 \sym{API}{Application Programming Interface}
 \sym{BERT}{Bidirectional Encoder Representations from Transformers}
 \sym{CV}{Cross-Validation}
 \sym{EEC}{Equity Evaluation Corpus}
 \sym{FKN}{Flesch-Kincaid Grade Level}
 \sym{FRES}{Flesch Reading-ease score}
 \sym{GCP}{Google Cloud Platform}
\sym{GLUE}{General Language Understanding Evaluation}
\sym{GPT}{Generative Pre-trained Transformer}
\sym{GRU}{Gated Recuurent Units}
\sym{HMM}{Hidden Markov Models}
\sym{IMDB}{Internet Movie Database}
\sym{IOB}{Inside, Outside, Beginning}
\sym{LLM}{Large Language Model}
 \sym{LM}{Language Models}
 \sym{LSTM}{Long Short Term Memory}
 \sym{MLM}{Masked Language Modelling}
 \sym{NER}{Named Entity Recognition}
 \sym{NLTK}{Natural Language Toolkit}
 \sym{NLU}{Natural Language Understanding}
 \sym{NSP}{Next Sentence Prediction}
\sym{OPT}{Open Pre-trained Transformer}
 \sym{POS}{Parts of Speech}
 \sym{PPO}{Proximal Policy Optimization}
 \sym{QA}{Question Answering}
 \sym{RNN}{Recurrent Neural Networks}
 \sym{RoBERTa}{Robustly Optimized BERT}
 \sym{RQ}{Research Question}
 \sym{SQuAD}{Stanford Question Answering Dataset}
 \sym{ULMFit}{Universal Language Model Fine-tuning
for Text Classification}
\sym{VADER}{Valence Aware Dictionary and sEntiment Reasoner}
\sym{WAE}{White-Aligned English}

\end{abbreviations}

\pagenumbering{arabic}
\chapter{INTRODUCTION}
\label{chapter:introduction}

\noindent Language modelling is a well-known task in which the goal is to train a model to learn the joint probability of a given sequence of words (Bengio, 2000) \cite{bengio2000neural}. These models are referred to as \textit{“Language Models”} (LM) and are used to generate sequences of words that form sentences or paragraphs. Various architectures have been applied to this task, including Hidden Markov Models (HMM), Recurrent Neural Networks (RNNs) \cite{rnn}, Long Short-Term Memory Networks (LSTMs) \cite{lstm}, Gated Recurrent Units (GRU) (Cho et al., 2014) \cite{cho-etal-2014-properties}, ELMo \cite{peters-etal-2018-deep}, Universal Language Model Fine-tuning for Text Classification (ULMFiT) \cite{howard-ruder-2018-universal}, and more recently, transformer networks such as Bidirectional Encoder Representations from Transformers (BERT) \cite{bert} and Generative Pre-trained Transformer (GPT)\cite{gpt-3}.

\begin{figure}
\scalebox{0.4}{
    \centering
    \includegraphics{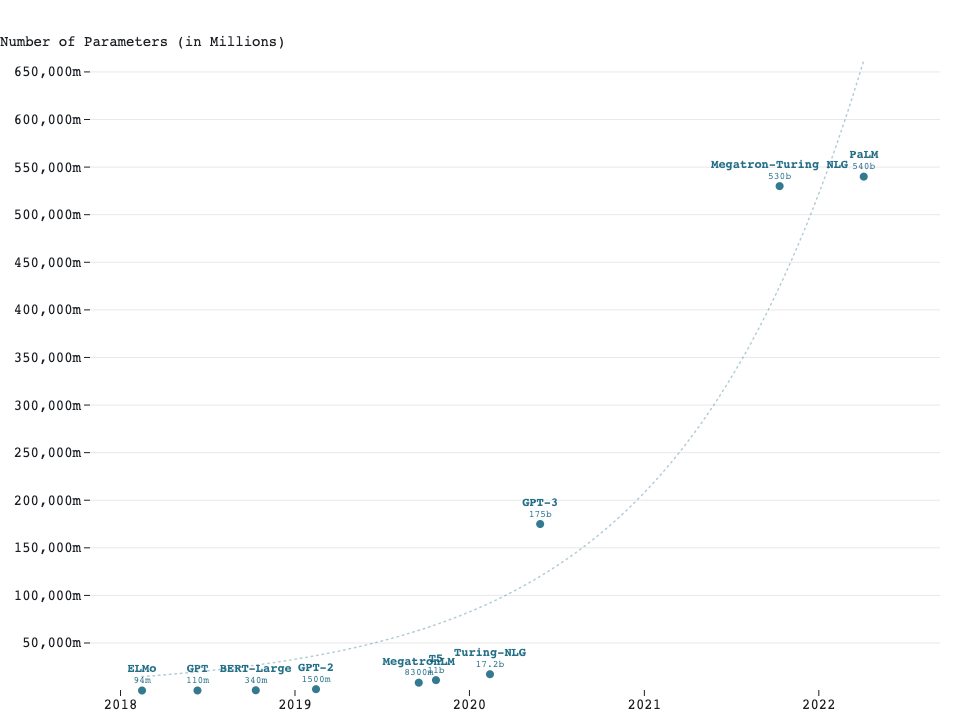}}
    \caption{Size comparison of Language Models.}
    \label{fig:lm_size}
\end{figure}

Significant progress has been made in the development of language models, as demonstrated by their performance on various benchmark datasets, such as the General Language Understanding Evaluation (GLUE) \cite{wang2018glue} and the SuperGLUE \cite{wang2019superglue}. The GLUE benchmark evaluates language models performance on several natural language understanding (NLU) tasks, including sentiment analysis, textual entailment, and question-answering. The highest-performing models on this benchmark have surpassed human-level performance on several tasks, although not all tasks have been surpassed. Much of this progress has been attributed to the idea of pre-training a general language model and further fine-tuning it on different sets of tasks. This approach has shown remarkable results, achieving state-of-the-art performance in various NLP and NLU tasks (\cite{bert}\cite{howard-ruder-2018-universal}). Over the years, different sizes of architectures have been developed, and while the performance of the models has improved, the size of the architectures has grown from a few hundred parameters to billions of parameters (see Figure \ref{fig:lm_size}). These large language models (LLM) are exemplified by GPT-3, which consists of 175 billion parameters and was trained on 570GB of text data \cite{gpt-3}.

The ability of Large Language Models (LLMs) to generate articles, essays, jokes, and even poetry in response to user inputs has opened up unique opportunities for the development of engaging applications. One recent advancement in this field is ChatGPT\footnote{Appication refer; https://openai.com/blog/chatgpt/}, a language model created by OpenAI \cite{chatgpt}. ChatGPT is specifically trained to interact with users in a conversational manner, answering follow-up questions, acknowledging its mistakes, challenging incorrect assumptions, and refusing inappropriate requests. The adoption and usage of ChatGPT have been substantial, making it the fastest-growing consumer application in history. In just two months, it has attracted over 100 million users\footnote{Available online and accessed 2nd Feb 2023 refer; https://www.reuters.com/technology/chatgpt-sets-record-fastest-growing-user-base-analyst-note-2023-02-01/}. This rapid growth highlights the widespread interest and acceptance of such conversational language models among users.

The deployment of large language models for public usage on such a scale does indeed come with inherent risks and challenges. One major concern is the presence of biases within these models. While LLMs are trained on vast amounts of filtered text data from the internet, this data may not be representative of major minority groups \cite{bender2021dangers}. Studies have shown that internet access and data contributions are more prevalent among younger individuals and those from developed countries (Pew Research Center)\footnote{Available online and accessed on April 2021 refer; https://www.pewresearch.org/internet/fact-sheet/internet-broadband/}. Additionally, platforms like Reddit have a higher representation of male users compared to female users (Pew Research Center, 2016)\footnote{The report is available online and was accessed on 25th Feb 2016 refer; https://www.pewresearch.org/journalism/2016/02/25/reddit-news-users-more-likely-to-be-male-young-and-digital-in-their-news-preferences/}. As a result, biases present in the training data may be amplified or perpetuated through LLMs.

Another significant challenge is the potential for LLMs to generate coherent yet factually incorrect text  \cite{xiao2021hallucination}. This can contribute to the dissemination of misinformation and fake news, as the models may produce fabricated reviews, comments, or news articles. Privacy concerns are also raised by the use of LLMs. These models can extract sensitive information from their training data, including personal information, passwords, and financial data (\cite{carlini2021extracting} \cite{pan2020privacy} ). This poses privacy risks if not adequately addressed. Lastly, LLMs can be susceptible to various attacks, such as adversarial attacks. These attacks aim to manipulate the output generated by the models, potentially leading to undesirable consequences.

Addressing these risks and challenges is crucial for the responsible deployment and usage of large language models in order to mitigate biases, combat misinformation, ensure privacy protection, and enhance the security of these systems.  It is also important to note that the potential harm extends not only to human society, which absorbs the text generated by these models but also to future language models if their training data includes the artificial text which is currently unexplored. Thus, this thesis aims to comprehend the impact of artificial text during the pre-training of language models.

\section{Research Objectives }
\noindent The ease of access to applications based on large language models (LLM) increases the risk of the proliferation of artificial data (text from another language model) on the internet, which is the primary source of training such language models. The inclusion of artificial data in the training of language models, sourced primarily from the internet, presents potential concerns. Two main issues arise in relation to this artificial data:

\begin{enumerate}
    \item Firstly, the quality of artificial data may be inferior to human-generated content, which can impact the performance of the models if such data is included during their training.
    
    \item Secondly, language models have the capability to learn biases and adopt abusive language patterns from the training data. Consequently, there is a risk that harmful ideologies, such as racism, sexism, and ableism, may be reinforced if the training data includes artificial content.
    
\end{enumerate}

\noindent While these issues are acknowledged, they have not been explicitly proven. Moreover, the increasing scale of data poses challenges in reliably detecting and excluding artificial text during future language model training. Without robust tools to confidently differentiate artificial text from human-generated text, the presence of artificial data is likely to persist.

In light of these concerns, this thesis aims to contribute by exploring the impact of artificial data on the pre-training of language models. The objective is to gain a deeper understanding of how artificial data influences these models during the pre-training phase.

\section{Research Questions}
\noindent \textit{\textbf{Research Problem}}: \textit{What is the effect of pre-training RoBERTa using the news articles written by ChatGPT on the performance compared to the model pre-trained on human written news articles?}

\noindent In order to tackle this research problem, we pose the following research questions (RQ):
\begin{enumerate}
    \item \textbf{Research Questions 1}: \textit{Does the RoBERTa model pre-trained using ChatGPT generated text have statistically inferior performance in terms of:}
    \begin{enumerate}
        \item \textbf{sub-question a}: \textit{Accuracy on Sentiment Classification task?}
        \item \textbf{sub-question b}: \textit{F1-score Named Entity Recognition task?}
        \item  \textbf{sub-question c}: \textit{F1-score on Question Answering task?}
    \end{enumerate}
    
    \item \textbf{Research Question 2}: \textit{Is the RoBERTa model pre-trained using ChatGPT generated text more biased towards certain demographics such as gender for sentiment classification task in terms of mean polarity difference?}
\end{enumerate}

\section{Relation to AI Program}

\noindent The field of AI is currently witnessing significant research advancements in language models, and large language models (LLMs) are at the forefront of this development. In line with this trend, our research addresses a future-oriented problem within the domain of LLMs, thereby contributing to the broader understanding of these systems in the field of AI. Given the relevance and significance of our research to the AI program at Utrecht University, it holds substantial relevance in the context of ongoing academic endeavours in this field.

\section{Research Outline}
\noindent The remainder of the research proposal is structured as follows, Chapter 2 begins by providing a theoretical background on different language models and the training process involved. It also reviews existing research on the detection of artificial text generated by language models and discusses the current state of research on identifying bias in language models. Chapter 3 focuses on the methodology employed in the study. It outlines the data generation process using ChatGPT and describes the methods used to compare the language of CNN journalists with that of ChatGPT. Additionally, it explains the process of pre-training the language model, along with the techniques used for fine-tuning and evaluating the trained models in terms of performance and bias metrics. Chapter 4 presents the results obtained from the experiments conducted in the study. Chapter 5 involves a comprehensive discussion of the results and explores the limitations encountered during the research and suggests potential directions for future research that can be pursued. The Chapter 6 addresses the ethical issues and environmental impacts associated with training language models in the study. Finally, we conclude our research questions and thesis in Chapter 7.

\chapter{BACKGROUND AND RELATED WORK}
\label{chapter:background}
\noindent In this chapter, we aim to provide a comprehensive overview of the prior work relevant to our research. We begin with Section 2.1, which presents the theoretical background on language models and outlines the training process for popular language models. Next, in Section 2.2, we delve into the existing research on detecting artificial text, exploring the various techniques and approaches employed in this area. Finally, in Section 2.3, we discuss the relevant studies and methodologies related to bias measurement in language models.

\section{Background on Language Models}
\noindent Language models have traditionally been trained using next-word prediction tasks, where the goal is to predict the next word given a sequence of words. This approach has been applied to various models, including Hidden Markov Models, RNNs \cite{rnn}, and LSTMs \cite{lstm}. However, with the introduction of the transformer architecture \cite{transformer}, a new training method called Masked Language Modelling (MLM) emerged \cite{bert}. In MLM tasks, certain random words in a sentence are masked, and the model is trained to predict the correct replacement words for these masks. This approach has been widely adopted by transformer-based models such as BERT \cite{bert}, RoBERTa \cite{roberta}, and ALBERT \cite{albert}. These models have demonstrated strong performance in a variety of natural language processing tasks. Additionally, there are transformer-based language models trained using next-word prediction, which are commonly referred to as auto-regressive models. Notable examples include XL-Net \cite{xl-net} and the GPT family \cite{gpt-3}. These models generate text by predicting the next word based on the previous context, allowing for the generation of coherent and contextually relevant sequences of words.

\textbf{Instruction fine-tuning}: Traditionally, language models have been trained using the objective of predicting the next word in a sequence of words. However, there has been a recent shift towards training language models to follow user instructions. This approach enables language models to perform a wide range of tasks, including summarization, translation, and classification, based on the instructions provided \cite{raffel2020exploring, chung2022scaling, ouyang2022training}.

The instruction-based training paradigm allows language models to exhibit more controlled behavior and generate output that aligns with user expectations. Instead of relying solely on context and statistical patterns, language models can now leverage explicit instructions to guide their generation process. This alignment of outputs is primarily achieved through the process of fine-tuning the language model on user instructions, incorporating reinforcement learning from human feedback \cite{stiennon2020learning, ouyang2022training}.

\begin{figure}
    \centering
    \scalebox{0.35}{
    \includegraphics{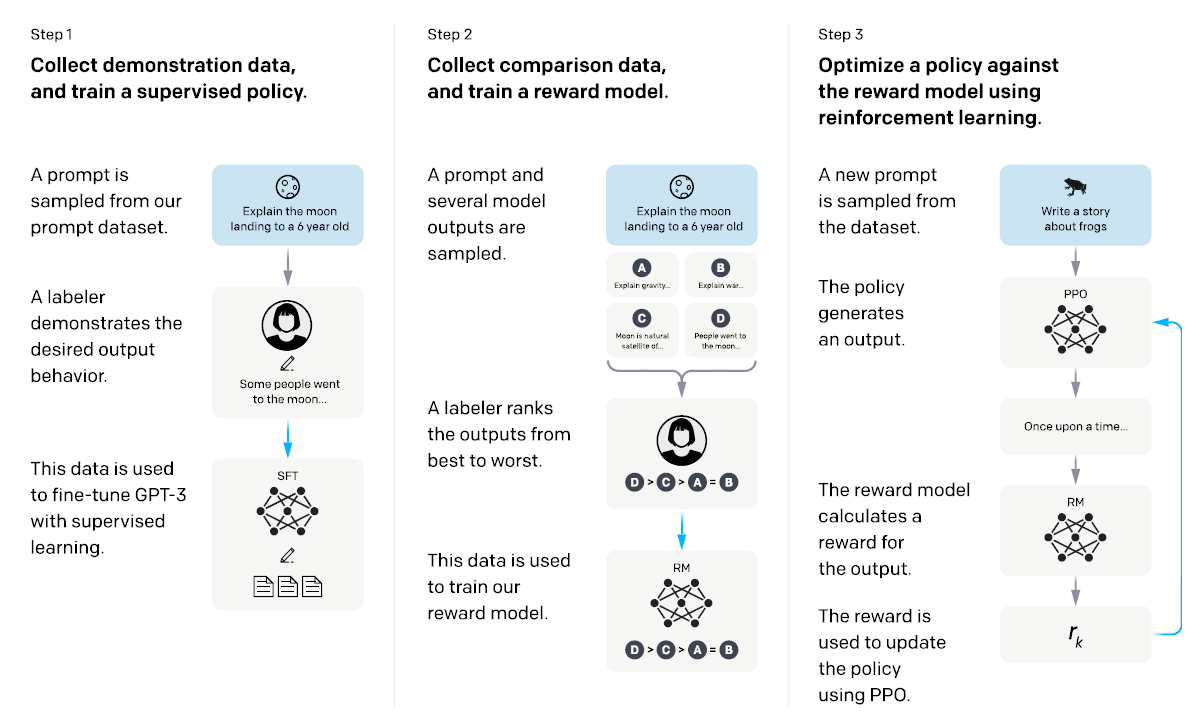}}
    \caption[Instruction fine-tuning for InstructGPT.]{Instruction fine-tuning for InstructGPT \cite{ouyang2022training}.}
    \label{fig:chatgpt-training}
\end{figure}

One notable example of an instruction-based language model is InstructGPT, which is a sibling model to ChatGPT. InstructGPT was trained to follow user instructions using a three-step process, as depicted in Figure \ref{fig:chatgpt-training}. The first step involved supervised fine-tuning using a collection of demonstration data annotated by human labellers. This data consisted of examples where human labellers provided instructions and the desired model response. In the second step, a reward model was trained based on human preferences for model output. This involved collecting human rankings or preferences for different outputs generated by the model, allowing the reward model to learn the desired output patterns. The third and final step focused on training the model to optimize its reward policy based on human preferences. Proximal Policy Optimization (PPO) \cite{schulman2017proximal} is used to fine-tune the model and improve its performance according to the reward policy learned in the previous step. While the complete details of the ChatGPT training process are not publicly available, it is trained using a similar methodology, as mentioned in the OpenAI blog\footnote{ChatGPT release refer; https://openai.com/blog/chatgpt}.

\subsection{Training Data for Language Models}
\noindent There is significant variation in the size of training data required for different architecture types, as summarized in Table \ref{tab:lang_model}. For instance, BERT primarily used the BookCorpus and Wikipedia datasets, which amounted to approximately 16GB of uncompressed text. RoBERTa, on the other hand, utilized five English-language corpora spanning various sizes and domains, totalling over 160GB of uncompressed text. XL-Net was trained on a compiled text dataset of 126GB, with further details provided in their paper \cite{xl-net}. The GPT-2 model relied on around 40GB of text data from sources such as Common Crawl and WebText, excluding Wikipedia articles specifically. Finally, the models in the GPT-3 family were trained using a vast dataset of 570GB of text data sourced from diverse data sources.

To train the InstructGPT model, distinct datasets were utilized for each step of the training process. The supervised fine-tuning dataset consisted of approximately 13 thousand training prompts, which were derived from a combination of prompts from the API and prompts generated by labellers. This dataset was used to fine-tune the model through a supervised learning approach. The reward model dataset, on the other hand, comprised around 33 thousand training prompts. These prompts were also sourced from a combination of prompts from the API and prompts generated by labellers. This dataset was employed to train a reward model, which provided feedback to the model based on human preferences for the generated outputs. Finally, for the Proximal Policy Optimization (PPO) step, a dataset consisting of approximately 31 thousand training prompts solely from the API was used. Further details about the dataset can be found in the original paper\cite{ouyang2022training}.


\begin{table}[]
    \centering
    \scalebox{0.8}{
    \begin{tabular}{|c|c|c|}
    \hline
        \textbf{Model} & \textbf{Data Source} & \textbf{Task} \\
        \hline
       BERT \cite{bert}& BooksCorpus, Wikipedia (Eng)& MLM and next-sentence prediction (NSP)\\ 
       \hline
       RoBERTa \cite{roberta}& BooksCorpus, Wikipedia (Eng) & MLM \\
        & CC-News, Stories, OpenWebText & \\
       \hline
       XL-Net \cite{xl-net}& BooksCorpus, Wikipedia (Eng),& Auto-regressive\\
       & Giga5, ClueWeb 2012-B,  Common Crawl & \\
       \hline
       GPT-2 \cite{radford2019language} & Common Crawl, WebText & Auto-regressive\\
       \hline
       GPT-3 \cite{gpt-3} & Common Crawl, WebText2 & Auto-regressive \\
       & Books1, Books2, Wikipedia & \\
       \hline
    \end{tabular}}
    \caption{Training details for popular language models.}
    \label{tab:lang_model}
\end{table}

\subsection{Overview of architecture}
\noindent In this section we provide a brief overview of the architecture design of some of the popular language models relevant in our study.

\begin{figure}
    \centering
    \scalebox{0.06}{
    \includegraphics{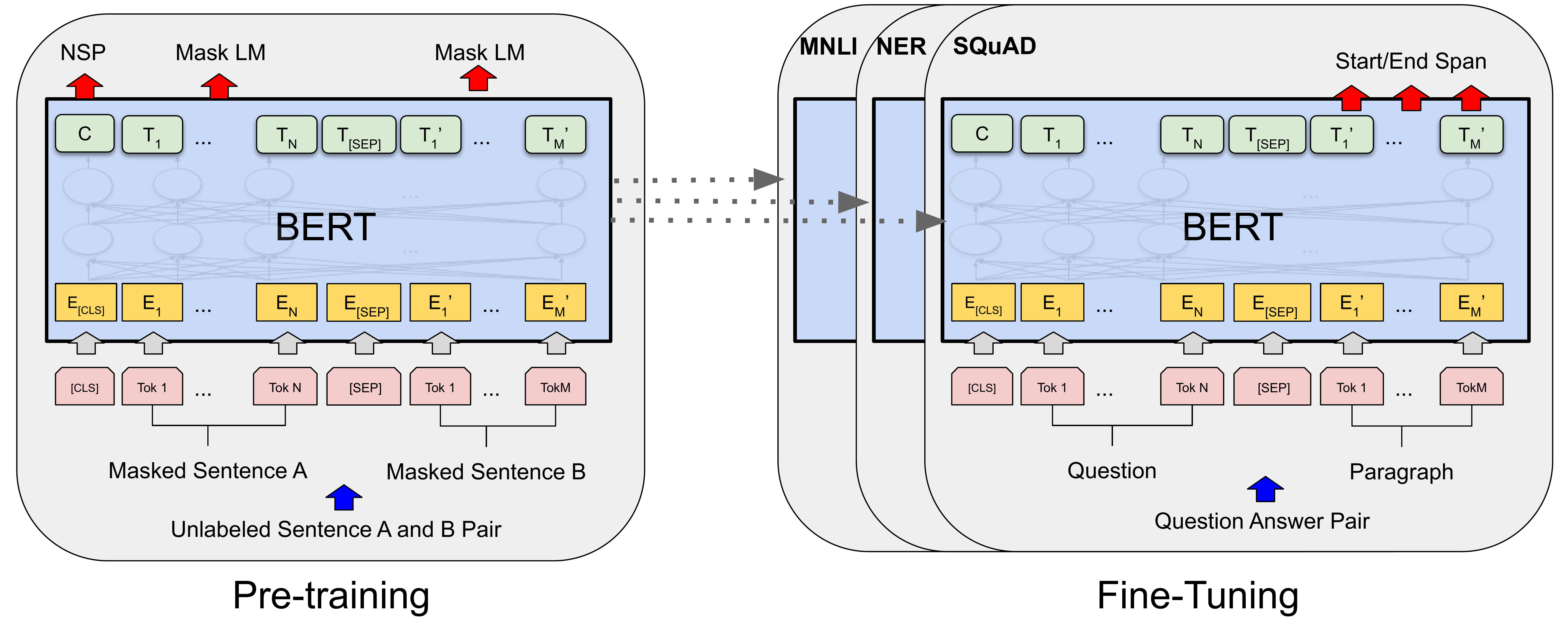}}
    \caption[Training process of BERT.]{Training process of BERT \cite{bert}.} 
    \label{fig:bert}
\end{figure}

\textbf{BERT}: BERT (Bidirectional Encoder Representations from Transformers) is a pre-trained language model that has gained significant attention in the field of natural language processing. It was developed based on the Transformer architecture, which was first introduced in the influential paper \textit{“Attention Is All You Need”} by Vaswani et al. \cite{transformer}. The BERT model is characterized by its encoder-only architecture and is available in two sizes: base and large. The base version of BERT consists of 12 layers, 768 hidden units, and 12 attention heads, resulting in a total of 110 million parameters. On the other hand, the large version of BERT is more extensive, with 24 layers, 1024 hidden units, and 16 attention heads, amounting to a total of 340 million parameters. During pre-training, BERT is trained with two main objectives: masked language modelling (MLM) and next sentence prediction (NSP). The MLM objective involves randomly masking a certain percentage of the input tokens and training the model to predict the masked tokens. The NSP objective, on the other hand, involves predicting whether two sentences appear consecutively in the training data as illustrated in Figure \ref{fig:bert}.


\textbf{RoBERTa}: Robustly Optimized BERT (RoBERTa) is a variant of BERT that is solely based on the Transformer encoder architecture. Similar to the BERT model, RoBERTa is available in two sizes: base and large. The base version of RoBERTa consists of 12 layers, 768 hidden units, and 12 attention heads, resulting in a total of 110 million parameters. On the other hand, the large version of RoBERTa is more extensive, with 24 layers, 1024 hidden units, and 16 attention heads, amounting to a total of 340 million parameters. During training, RoBERTa focuses solely on the masked language modelling (MLM) task. The training dataset used for RoBERTa comprises English-language corpora from various sources and domains, totalling over 160GB of uncompressed text. This diverse and extensive dataset contributes to the robustness and generalization capabilities of RoBERTa.


\textbf{GPT-2}: The GPT-2 (Generative Pre-trained Transformer 2) model, developed by OpenAI \cite{radford2019language}, is a large-scale natural language processing (NLP) model. It is pre-trained on an extensive corpus of text data, comprising 40 GB of text from various sources. The primary objective of pre-training GPT-2 is to develop a language model capable of predicting the next word in a sequence given the preceding context. The model is available in four different sizes, each varying in the number of layers, hidden dimensions, and total parameters. The smallest variant of GPT-2 consists of 12 layers and 768 hidden dimensions, resulting in a total of 117 million parameters. The medium version comprises 24 layers with 1,024 hidden dimensions, amounting to 345 million parameters. The large variant encompasses 36 layers with 1,280 hidden dimensions, yielding 762 million parameters. Lastly, the extra-large version boasts 48 layers with 1,600 hidden dimensions, resulting in about 1.5 billion parameters.

\textbf{GPT-3}: The GPT-3 (Generative Pre-trained Transformer 3) developed by OpenAI \cite{gpt-3} encompasses a family of eight models, each with varying numbers of parameters. These models range from the smallest model with 125 million parameters to the largest model with a staggering 175 billion parameters. The training methodology for the GPT-3 models is similar to that of GPT-2 but with some notable differences. The GPT-3 models were trained on an extensive dataset consisting of approximately 300 billion tokens. The GPT-3 family of models exhibits differences in architecture depth across its members. The smallest model in the family features 12 layers, while the medium, large, and XL models comprise 24 layers. The 2.7B and 6.5B models have 32 layers, the 13B model has 40 layers, and the largest model, with 175 billion parameters, has 96 layers. These models form the base of the InstructGPT model and therefore have the same architecture.

\subsection{How much data do you need for pre-training?}
\label{data-size}
\noindent 
The size of the training data plays a crucial role in the performance and capabilities of language models. Different architectures, such as BERT and GPT-2, have been trained on datasets of varying sizes. For example, BERT was pre-trained on a dataset of 16GB of text, while GPT-2 utilized a much larger dataset of 40GB of text. Notably, studies such as Perez-Mayos et al. \cite{perez-mayos-etal-2021-much} and Zhang et al. \cite{zhang-etal-2021-need} have specifically focused on understanding the impact of data size on model performance. Their findings contribute to our understanding of the trade-offs and considerations associated with training language models on different data sizes.



\begin{figure}[h!]
    \centering
    \includegraphics[scale=0.5]{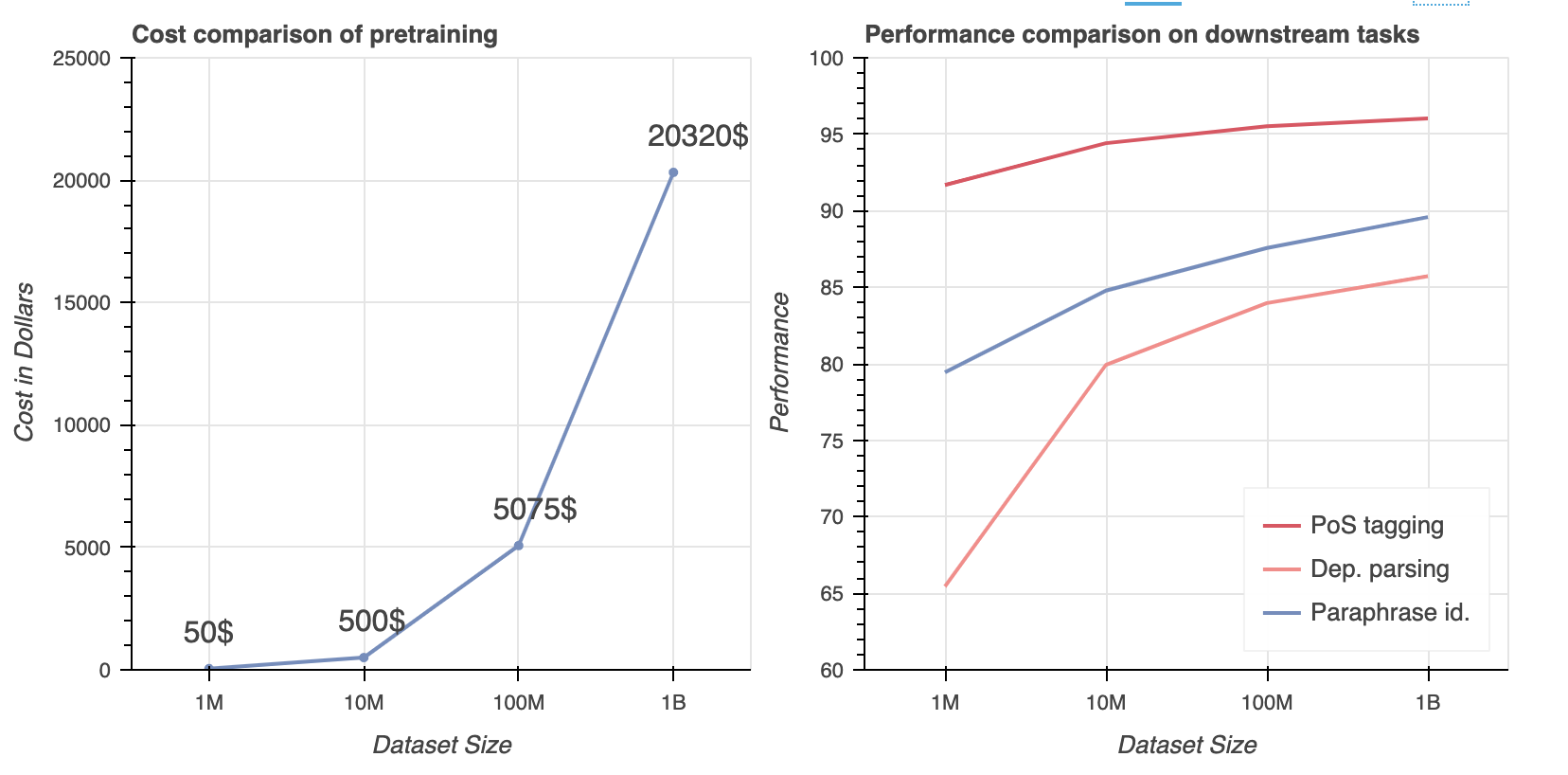}
    \caption[Cost vs Performance tradeoff for RoBERTa model]{Cost vs Performance tradeoff for RoBERTa model (data for the figure is taken from the study done by Perez et al. \cite{perez-mayos-etal-2021-much} )}
    \label{fig:cost_camp}
\end{figure}

In a study conducted by Perez et al. \cite{perez-mayos-etal-2021-much}, a cost-benefit trade-off was performed for a RoBERTa model trained on different-sized datasets, ranging from 1M to 1B tokens. The analysis, depicted in Figure \ref{fig:cost_camp}, showed that the performance gains were most significant when increasing the dataset size from 1M to 10M tokens, while the cost of pre-training increased substantially. Another study also found that the model was able to learn the syntax of the language effectively with approximately 10M tokens \cite{zhang-etal-2021-need}. Taking these findings into consideration, we aim to construct our dataset for pre-training our models in an efficient manner.

\section{Detection of Artificial text}
\noindent With the increasing prominence of Large Language Models (LLMs), there has been a surge in research efforts aimed at detecting text generated by these models. In this section, we present an overview of various techniques that have been explored for the purpose of detecting artificial text. However, it is important to note that these methods have inherent limitations and are not sufficient to fully address the challenge of identifying artificial text. We will discuss the shortcomings of these approaches and highlight the need for further advancements in this area to effectively tackle the problem of distinguishing between human-generated and LLM-generated text.

\subsection{Post-hoc Detection}
\noindent 
The post-hoc technique aims to develop models capable of detecting artificial text by leveraging features extracted from the language model or by fine-tuning language models for text classification \cite{zellers2019defending, tan2020detecting}. However, the generalizability of detectors trained for one language model to other models has not yet been established. This limitation was evident in a study conducted by Gambini et al. \cite{gambini2022pushing}, where detectors trained for GPT-2 struggled to accurately detect text generated by the GPT-3 model. Furthermore, these detectors are susceptible to adversarial attacks, as demonstrated by Wolff et al. \cite{ad-dec}. 

OpenAI recently made an attempt to train a classifier specifically designed to distinguish between text generated by humans and text generated by AI models from various providers \cite{gpt-detection}. However, the reliability of this classifier was found to be limited. During the evaluation, it only correctly identified 26\% of AI-written text as \textit{“likely AI-written”} (true positives), while it incorrectly classified human-written text as AI-written in 9\% of cases (false positives). OpenAI also acknowledged certain limitations of the classifier, such as lower reliability for shorter text and languages other than English. Additionally, there is a possibility of evading detection by manipulating or editing the text. 

In theory, there is a possibility to train language models using adversarial samples, which could potentially allow them to evade detection methods. As a result, relying solely on post-hoc detection techniques may not be a comprehensive solution for effectively identifying and mitigating the potential harm that could arise from large language models (LLMs). 

\subsection{Watermarking in Language Models}
\noindent Watermarking in text generation is a technique that introduces hidden signals into generated text, which can be used to detect whether the text is artificial. In this approach, the language model generates a probability distribution for the next word based on a given sequence of words. The next word is then randomly selected from this distribution. The watermarking technique operates by pseudo-randomly sampling words from the distribution, thereby enabling the detection of text generated by the language model.

In a recent study conducted by Kirchenbauer et al. \cite{kirchenbauer2023watermark}, the effectiveness of the watermarking technique was evaluated on a multi-billion parameter model from the Open Pre-trained Transformer (OPT) family. The proposed method involved the selection of randomized sets of \textit{green} tokens (allowed tokens) and \textit{red} tokens (restricted tokens) during the generation process. The use of green tokens was promoted while sampling, while red tokens were discouraged. This approach facilitated the detection of artificial text by counting the occurrences of red tokens. In a similar direction, OpenAI has also expressed intentions to incorporate watermarking into its language models. As outlined in a blog post by Scott Aaronson \footnote{Blog refer; https://scottaaronson.blog/?p=6823}, OpenAI plans to use cryptographic pseudo-random functions to sample words and generate detectors using cryptographic keys. This approach aims to embed watermarks into the language model, allowing for the identification of artificial text.

In practice, evading watermarking detection in language models can be achieved through various techniques, such as inserting or deleting words randomly, rearranging sentence order, or paraphrasing text. These methods pose challenges to the effectiveness of watermarking as a detection mechanism. Additionally, the responsibility of embedding watermarking lies with the developer of the language model, and only they have the ability to build the corresponding detection models. This limitation means that watermarking can only mitigate the potential harms of language models to a certain extent, as it relies on the actions and choices of the developer.

\section{Bias Measures}
\label{bias-literature}
\noindent In the realm of decision-making AI, the term bias refers to the \textit{act of making judgments or classifications based on preconceived notions or prejudices rather than impartially evaluating the available facts} \cite{kate-bias}. Traditionally, the evaluation of bias in systems has relied on well-established approaches such as equalized odds, equal opportunity, and demographic parity \cite{dwork2012fairness, hardt2016equality}. However, these methods are primarily suited for assessing bias in decision-making systems and may not directly apply to generative applications of language models (LLMs), which encompass user interaction and open-ended text generation. The evaluation of bias in generative text generation is an active area of research, with efforts aimed at studying various types of biases, including but not limited to gender, racial, cultural, and political or ideological biases. Researchers have explored different approaches to uncover and measure these biases within the context of LLMs. These approaches are tailored to address the unique challenges posed by generative text generation and enable a deeper understanding of the biases that may arise. In the following sections, we provide an overview of some of these approaches that have been employed in the study of bias in LLMs. These approaches offer valuable insights into the nature of biases that can emerge in generative text generation and contribute to the ongoing efforts to develop fair and unbiased AI systems.

\subsection{Measuring Bias in a Natural Language Generation Task}
\noindent In the field of natural language generation, Sheng et al. \cite{sheng-etal-2021-societal} have provided a comprehensive review of the literature on examining societal biases in the output generated by natural language models. Several notable works in this area include:

 \textbf{Regard Ratio}: Sheng et al. (2019) \cite{sheng-etal-2019-woman} introduced the concept of \textit{regard} as a metric for evaluating bias in natural language generation. They conducted an analysis of text generated from prompts containing mentions of diverse demographic groups. The goal was to measure the sentiment towards these groups and identify potential biases. To accomplish this, they proposed the development of a regard classifier.

 \textbf{Gendered Word Co-occurrence Score}: Bordia et al. \cite{bordia-bowman-2019-identifying} proposed a methodology to define bias in natural language generation, based on the absolute log ratio of probabilities between $P(word | female-terms)$ and $P(word | male-terms)$ for all words in the generated text. This approach allows for the systematic identification of differences in the usage of female and male terms, which can indicate the presence of gender bias. By utilizing statistical measures to quantify the extent of bias, Bordia and Bowman were able to evaluate and compare the levels of bias across different language models and datasets. 

 \textbf{Sentiment Ratio}: Groenwold et al. \cite{groenwold-etal-2020-investigating} also studied racial bias in natural language generation by measuring the sentiment score ratios of text generated from African American English (AAE) versus White-Aligned English (WAE) prompts. Their work sought to identify differences in the sentiment scores of the generated text that may be attributed to racial biases.

\subsection{Measuring Bias in a Classification Task}

\noindent In the domain of sentiment analysis, Kiritchenko et al. \cite{sentiment-bias} developed the Equity Evaluation Corpus (EEC), which comprises 8,640 English sentences aimed at assessing biases related to specific races and genders. The EEC has been utilized by various studies to investigate bias in different contexts. For instance, Bhardwaj et al. \cite{bhardwaj} examined gender bias in BERT embeddings using the EEC, while Huang et al. \cite{counter-bias} evaluated bias through the application of fairness metrics, both at the group and individual levels, by manipulating the context of the samples using counterfactuals. In the downstream application within the clinical domain, Gizem et al. \cite{sogancioglu2022gender} demonstrated the existence of societal bias in the contextualized embeddings.

In another study, Jentzsch et al. \cite{jentzsch-turan-2022-gender} conducted an analysis of gender bias in sentiment classification using the IMDB dataset. They introduced male and female versions of each sample for evaluation purposes and defined gender bias as the mean difference in polarity scores across all samples. Additionally, they performed fine-tuning experiments with different training data to investigate the individual impacts of pre-training and fine-tuning. This involved comparing models trained on the original dataset to models trained after removing gender-specific words from the training dataset or creating a balanced dataset with male and female versions of each sample. These investigations shed light on the influence of pre-training and fine-tuning on gender bias in sentiment analysis models.

\chapter{METHODOLOGY}
\label{chapter:proposedmethod}
\noindent In this chapter, we will provide a comprehensive overview of the dataset utilized for the pre-training phase as well as for the downstream tasks. We will detail the methodology employed for training the models and outline the approach taken for evaluating their performance.

\section{Dataset}
\subsection{CNN/DailyMail Dataset}

\noindent In order to pre-train the RoBERTa language model, a large text corpus, is necessary. The original RoBERTa pre-training process incorporated a combination of diverse sources such as BooksCorpus, English Wikipedia, CC-News, Stories, and OpenWebText ( for masked language modelling, MLM task ). For our research, we selected the \textbf{CNN/DailyMail Dataset} as the primary source of text data for pretraining RoBERTa. This choice enables us to compare the content style and size between human-written articles from CNN and the Daily Mail, while also focusing on a practical task relevant to the application of ChatGPT or other large language models in the future.

The CNN/DailyMail Dataset is an extensive collection of English-language news articles, consisting of approximately 300,000 distinct articles authored by journalists from CNN and the Daily Mail. To facilitate the pretraining of language models, a subset of around 25,000 articles is sampled from the dataset and divided into separate training and evaluation sets. To support accessibility and promote further research, we have made this dataset openly available through a public repository hosted by Hugging Face\footnote{Dataset available for download; https://huggingface.co/datasets/isarth/chatgpt-news-articles}.
 
\begin{figure}
    \centering
    \scalebox{1.5}{
    \includegraphics{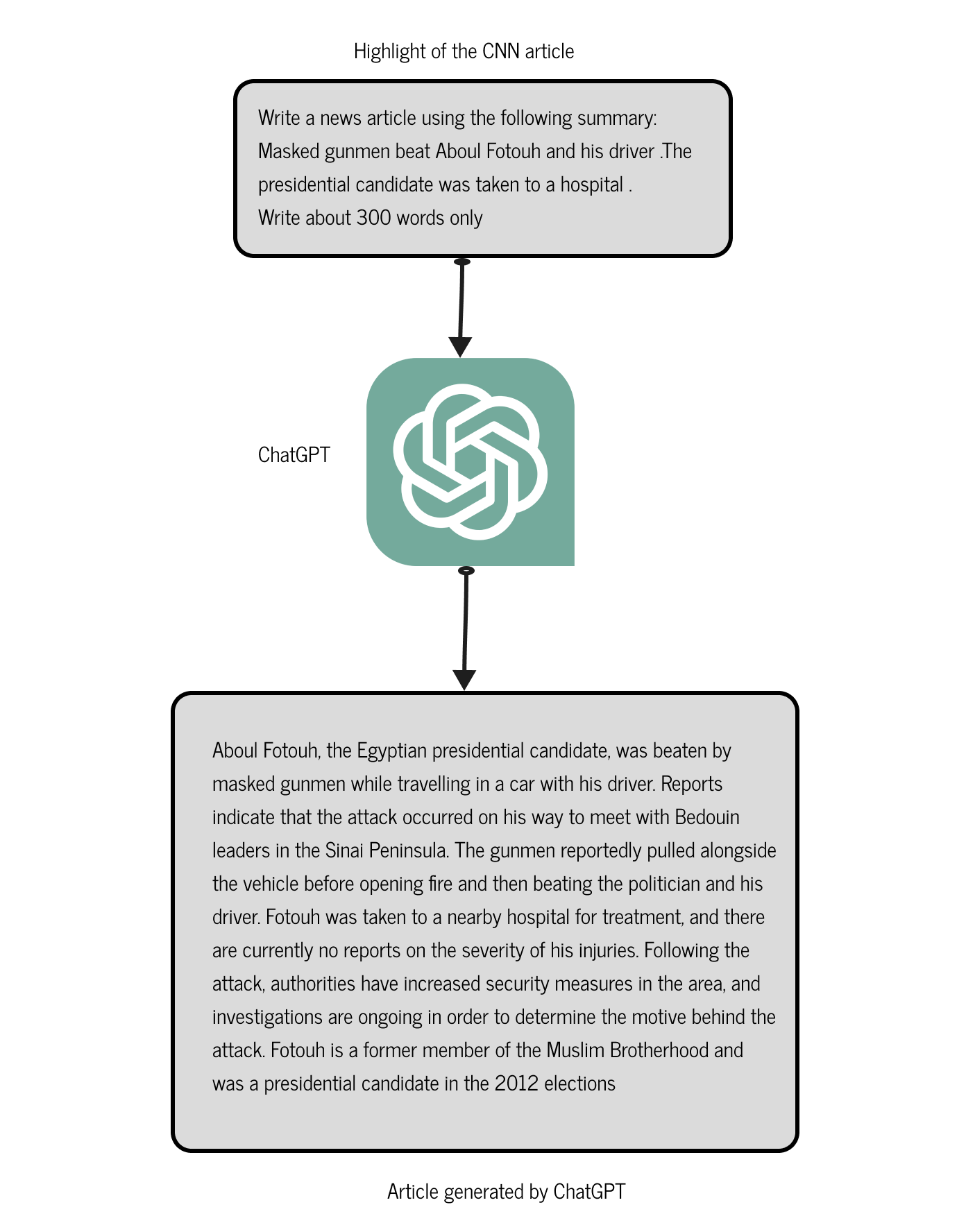}}
    \caption{Process of generating news articles using ChatGPT.}
    \label{fig:chatgpt_art}
\end{figure}

\subsection{News Article Generation from ChatGPT}
\noindent  Each article within the CNN/DailyMail dataset is accompanied by a human-generated abstractive summary. These summaries serve as prompts for ChatGPT to generate news articles. By leveraging this information, ChatGPT is able to produce articles in response to specific prompts or topics. To instruct ChatGPT, we used the prompt \textit{You are an AI assistant that generates news articles from a summary}. No prompt engineering was performed, and we aimed to keep the prompt for  generation as simple as possible. 

During the process, the word count of the original article is calculated, and ChatGPT is directed to generate an article with the same number of words. This ensures consistency in the length of the generated articles. The pseudo-code below provides the precise prompt used for the ChatGPT API, and Figure \ref{fig:chatgpt_art} illustrates an example of a generated article.

\begin{algorithm}
\caption{ChatGPT Prompt}
Model = gpt-3.5-turbo \\
Role[System]= You are an AI assistant that generates news articles from a summary.
Role[User] = Write a news article using the following summary: \\
. \hspace{2cm}  [Highlights of the original article] \\
. \hspace{2cm} Write about [Num of words in the original article] words only.
\end{algorithm}

\noindent All articles were compiled and generated in April 2023 using the 0301 version of the ChatGPT-turbo API. The cost of generating 25,000 articles was approximately 31 dollars. However, information regarding the carbon compute and the specific server used by OpenAI is currently unavailable, so we cannot provide details about the carbon emissions associated with the generation process.

\section{Qualitative Analysis of Dataset}
\noindent 

\noindent Understanding the differences in style and content between human writing and the output of ChatGPT is a crucial aspect of our study. It allows us to examine the variations in performance between the RoBERTa architecture during pre-training and throughout different downstream tasks. To conduct a comparative analysis of the article corpus generated by CNN/DailyMail journalists and ChatGPT, several key statistical measures were computed. These measures encompass the total word count, average word count per article, total vocabulary size, average number of sentences per article, average word count per sentence, and average number of named entities per article. Named entities were identified using the Spacy library \cite{spacy2}, which provides a comprehensive suite of natural language processing tools.

Furthermore, we analyze the sentiment intensity of each article using VADER (Valence Aware Dictionary and sEntiment Reasoner) \cite{vader}, a sentiment analysis tool available in the Natural Language Toolkit (NLTK) library \cite{nltk}. VADER assigns compound sentiment scores between 0 and 1 for each sentiment category (positive, negative, and neutral). For example, an article may have compound scores of 0.8 for positive sentiment, 0.15 for negative sentiment, and 0.05 for neutral sentiment. This analysis provides insights into the overall sentiment expressed in the generated articles.

\subsection{Readability Metrics}
\noindent We utilized two widely used readability metrics, namely the Flesch-Reading Ease and Flesch-Kincaid Grade Level, to assess the readability of the generated articles.

The \textbf{Flesch-Reading Ease score} \cite{flesch} is a measure commonly employed to indicate the difficulty level of a passage. It is based on two key factors: sentence length and word length, with some weighting. Lower Flesch-Reading Ease scores indicate that the text or passage is more challenging to read and requires a higher level of education to understand. Conversely, higher scores indicate that the text or passage is easier to read and comprehend, requiring a lower level of education. Table \ref{tab:flesch_inter} provides an interpretation of the Flesch-Reading Ease scores. The formula for calculating the Flesch-Reading Ease score (FRES) is given in Equation \ref{eq:fres}.
\begin{equation}
\label{eq:fres}
 FRES(text) = 206.835 - 1.015(\frac{total\_words}{total\_sentences}) - 84.6(\frac{total\_syllables}{total\_words}) 
 \end{equation}
 
 The \textbf{Flesch-Kincaid Grade Level} \cite{flesch} is another widely used readability metric, particularly in the field of education. It provides a numerical score that corresponds to a U.S. grade level. This metric can give an estimate of the number of years of education typically required to understand the text. The Flesch-Kincaid Grade Level Formula (FKN) is used to calculate the grade level, as shown in Equation \ref{eq:fkn}.
\begin{equation}
\label{eq:fkn}
   FKN(text) =  0.39(\frac{total\_words}{total\_sentences}) + 11.8(\frac{total\_syllables}{total\_words}) - 15.59
\end{equation}

\begin{table}[]
    \centering
    \begin{tabular}{ccr}
       \textbf{Score}  & \textbf{School level} & \textbf{Interpretation}\\
    \hline
        100-90 & 5th Grade & Very Easy \\
        90-80 & 6th Grade & Easy \\
        80-70 & 7th Grade & Fairly Easy \\
        70-60 & 8th \& 9th Grade & Standard \\
        60-50 & 10th to 12th Grade & Fairly Difficult \\
        50-30 & College & Difficult\\
        30-10 & College Graduate & Very Difficult \\
        10-0& Professional & Extremely Difficult/Confusing \\
    \hline
    \end{tabular}
    \caption{Interpretation of the scores of Flesch-Reading Ease.}
    \label{tab:flesch_inter}
\end{table}

\section{Language Modeling and Pre-training}

\noindent The main objective of pre-training the RoBERTa model is Masked Language Modeling (MLM). In our study, we conducted pre-training using two different versions of the RoBERTa architecture, with each model trained on a distinct text corpus. One model was pre-trained exclusively on a dataset composed of articles authored by CNN and Daily Mail journalists, while the other model was pre-trained using a dataset consisting of articles generated by the ChatGPT language model. By training one model on articles from reputable news sources like CNN and Daily Mail, we aimed to capture the stylistic and structural nuances characteristic of professional journalism. On the other hand, pre-training the second model on articles generated by ChatGPT allowed us to harness the conversational and generative abilities of the language model itself.

To ensure a fair comparison, we maintained identical parameters for pre-training both models. This allowed us to directly compare the models' performance and characteristics in terms of their language generation capabilities. In the pre-processing stage, we converted the entire text to lowercase for both corpora and pre-train uncased language models.

To provide transparency and reproducibility, we included all the relevant parameters used in the pre-training process of the RoBERTa models in the appendix. These parameters cover various aspects of the pre-training methodology, including hyper-parameters, model architecture configurations, training data specifications, and any additional settings considered significant for the experiments.

\section{Evaluating Downstream Performances}
\noindent To evaluate and compare the quality of the language models, we conducted a comprehensive analysis of their performance on a range of downstream tasks. We carefully selected a diverse set of tasks to cover various domains and challenges. The pre-trained language models were then fine-tuned on each specific downstream task using carefully selected datasets. 

\begin{figure}
    \centering
    \scalebox{0.7}{
    \includegraphics{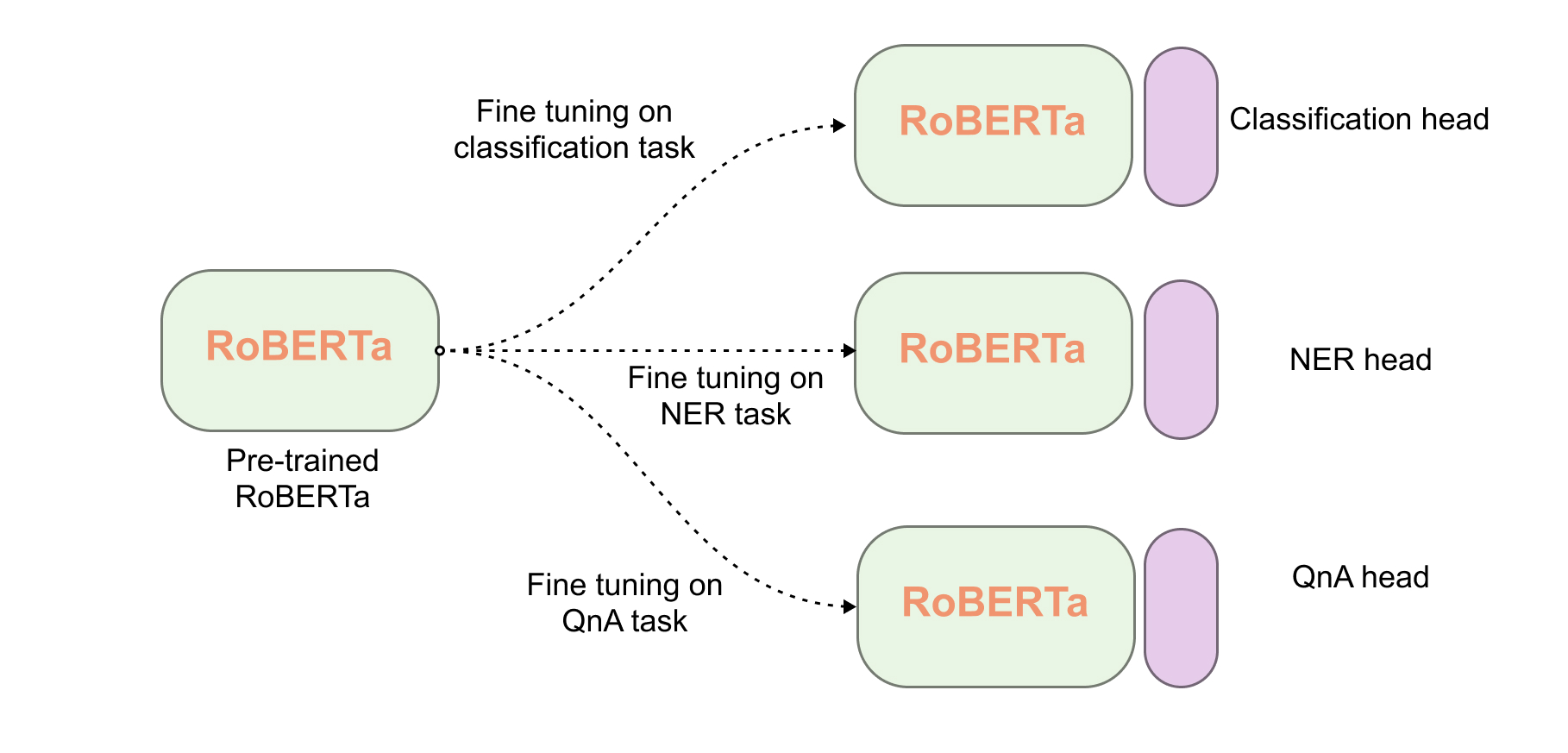}}
    \caption{Fine-tuning RoBERTa on different downstream tasks.}
    \label{fig:method_fine}
\end{figure}

During the fine-tuning process, we added task-specific layers and adjusted the weights and parameters of the models to optimize their performance for each task. We benchmark the results of the fine-tuned models against the original RoBERTa model that was fine-tuned for each downstream task. This allowed for a thorough evaluation of the effectiveness of the language models. To optimize the performance of the models, we performed hyper-parameter search using a small set of samples from the training dataset. Further details about hyper-parameters are mentioned in the Appendix Section \ref{parameters} and our code is also available on the Github\footnote{For the repository refer; https://github.com/sarthusarth/lang\_mod\_chatgpt}. In order to account for potential performance variance due to randomness \cite{reimers-gurevych-2017-reporting}, we employed 5x2 cross-validation to compare the trained models. Figure~\ref{fig:method_fine} provides an overview of the fine-tuning process for the different downstream tasks.

In the subsequent subsections, detailed information regarding each downstream task is provided. This includes task-specific datasets, evaluation metrics, and any specific considerations or challenges associated with the task. 

\subsection{Sequence Classification}
\begin{table}[ht]
    \centering
    \begin{tabular}{|p{0.4\linewidth} | p{0.4\linewidth}|}
      \hline
      \textbf{Positive Sample}  & \textbf{Negative Sample} \\ \hline
      
      i m a male not given to women s movies but this is really a well done special story i have no personal love for jane fonda as a person but she does one hell of a fine job while deniro is his usual superb self everything is so well done acting directing visuals settings photography casting if you can enjoy a story of real people and real love   this is a winner & this fanciful horror flick has vincent price playing a mad magician that realizes his vocational talents have been sold to another he devise ways of avenging all those that have wronged him his master scheme seems to back fire on him price is a little below par compared to his masterpieces but is still the only reason to watch this thriller supporting cast includes patrick o neal mary murphy eva gabor and jay novello \\
      \hline
    \end{tabular}
    \caption{Samples from the IMDB dataset.}
    \label{tab:imdb_sample}
\end{table}

\noindent For the sequence classification task, we evaluated the models on the IMDB sentiment challenge \cite{maas-EtAl:2011:ACL-HLT2011}. The goal of this task is to classify movie reviews into positive and negative sentiments. The original dataset contains 25,000 training samples and 25,000 test samples.

During the pre-processing step, we lowercase the dataset since the models were pre-trained on an uncased text corpus. Table \ref{tab:imdb_sample} displays examples from the pre-processed dataset.

To train the models, we use 20\% of the samples from the original training dataset for hyper-parameter search. Subsequently, we combine all the samples from the training and test sets to create splits for 5x2 cross-validation. This allows us to perform a comprehensive comparison of the models performance

\subsection{Named Entity Recognition (NER)}
\noindent For the Named Entity Recognition task, we evaluated the models on the WNUT 17: Emerging and Rare entity recognition challenge \cite{derczynski-etal-2017-results}. This task focuses on identifying unusual and previously unseen entities in the context of emerging discussions. The original dataset consists of 3,394 sentences for training and 1,287 sentences for testing. The dataset provides 12 possible Named Entity Recognition (NER) tags in Inside-Outside-Begin (IOB) format, as shown in Table \ref{tab:wnut_tags}.

\begin{table}[]
    \centering
    \scalebox{0.8}{
    \begin{tabular}{|c|c|c|}
    \hline
       \textbf{Number}  & \textbf{NER Tag} & \textbf{Description of Entity Tag} \\
       \hline
        0& O & Outside \\
1& B-corporation & Beginning of corporation \\
2& I-corporation & Inside of corporation \\
3& B-creative-work & Beginning of creative-work (song, movie, book and
so on)\\
4& I-creative-work & Inside of creative-work \\
5& B-group & Beginning of a group (sports team or non-corporate teams)\\
6& I-group & Inside of a group \\
7& B-location & Beginning of location (including GPE, facility) \\
8& I-location & Inside of location\\
9& B-person & Beginning of a person\\ 
10& I-person & Inside of a person \\
11& B-product & Beginning of a product (tangible goods, or well-defined
services) \\
12& I-product & Inside of a product\\
\hline
    \end{tabular}}
    \caption[NER tags  in the WNUT 17 dataset.]{NER tags (provided in IOB format) in the WNUT 17: Emerging and Rare entity recognition challenge.}
    \label{tab:wnut_tags}
\end{table}

During the training of the models, we use 20\% of the samples from the original training dataset for hyper-parameter search. Subsequently, we combine all the samples from the training and test sets to create splits for 5x2 cross-validation. This allows us to perform a comprehensive comparison of the models' performance.

\subsection{Question-Answering}
\noindent For the Question Answering task, we evaluated the models using the Stanford Question Answering Dataset (SQuAD) \cite{rajpurkar-etal-2016-squad}. This dataset is designed for reading comprehension and consists of questions posed by crowd-workers on a set of Wikipedia articles. Each question has a corresponding reading passage, and the answer to each question is a segment of text, or span, from the passage. The original dataset includes 87,000 training questions and 10,000 validation questions. Each data point in the dataset contains a context, a question, and an answer, as shown in Table \ref{tab:squad_sample}.

During training, the models are provided with the context and question, separated by a token, and the goal is to predict the correct answer. For hyperparameter search, we use 20\% of the samples from the original training dataset. To compare the models, we combine all the samples from the training and test sets and create splits for 5x2 cross-validation. This enables a comprehensive evaluation of the models' performance. The original dataset can be accessed publicly from the Hugging Face repository.\footnote{https://huggingface.co/datasets/squad} 

\begin{table}[ht]
    \centering
    \begin{tabular}{|p{0.5\linewidth} | p{0.2\linewidth}|p{0.2\linewidth}|}
      \hline
      \textbf{Context}  & \textbf{Question} & \textbf{Answer} \\ \hline
      Architecturally, the school has a Catholic character. Atop the Main Building's gold dome is a golden statue of the Virgin Mary. Immediately in front of the Main Building and facing it, is a copper statue of Christ with arms upraised with the legend "Venite Ad Me Omnes". Next to the Main Building is the Basilica of the Sacred Heart. Immediately behind the basilica is the Grotto, a Marian place of prayer and reflection. It is a replica of the grotto at Lourdes, France where the Virgin Mary reputedly appeared to Saint Bernadette Soubirous in 1858. At the end of the main drive (and in a direct line that connects through 3 statues and the Gold Dome), is a simple, modern stone statue of Mary & What sits on top of the Main Building at Notre Dame? & a golden statue of the Virgin Mary \\
      \hline
    \end{tabular}
    \caption{Sample data-point in the SQuAD dataset.}
    \label{tab:squad_sample}
\end{table}

\section{Paired T-Test for Model Comparisons}

\noindent The 5x2 CV paired t-test is a statistical procedure introduced by Dietterich \cite{dietterich1998approximate} to address limitations in other methods when comparing the performance of two models. The hypothesis in the test are:

\begin{itemize}
    \item \textbf{Null Hypothesis}: The performance of the two algorithms is not significantly different.
    \item \textbf{Alternate Hypothesis}: The performance of the two algorithms is significantly different. 
\end{itemize}
\noindent In this test, the training dataset is randomly split into two equal parts (50\% training and 50\% testing) five times. In each of the five iterations, two models are trained (model A and B) each using a 2-Fold CV and their performances are evaluated ($p_A^{(i)}$ and $p_B^{(i)}$, where $i \in \{1, 2\}$). The equations for an iteration j are as follows:
\begin{align}
    p_j^{(1)} = p_{jA}^{(1)} - p_{jB}^{(1)} \\
    p_j^{(2)} = p_{jA}^{(2)} - p_{jB}^{(2)} \\
    \Bar{p_j} = \frac{p_j^{(1)} + p_j^{(2)} }{2} \\
    s_j^2 = (p_j^{(1)} - \Bar{p_j})^2 + (p_j^{(2)} - \Bar{p_j})^2 
\end{align}

\noindent Finally, once all five iterations are computed, we can estimate the t-statistics as follows:
\begin{align}
    t-statistic = \frac{p_1^{(1)}}{\sqrt{(1/5) \sum_{j=1}^{5}s_j^2 }}
\end{align}
\noindent In the above equation $p_1^{(1)}$ is $p_1$ from the first iteration. The t-statistic is  assumed to approximately follow the t distribution with 5 degrees of freedom. Using the t statistic, the p-value can be computed and compared with a significance level of ($\alpha$ =0.05). If the p-value is lower than $\alpha$, we reject the null hypothesis and accept that there is a significant difference between the two models.

\section{Evaluating Biases}
Numerous studies have been conducted to evaluate bias in language models, as discussed in Section \ref{bias-literature}. In our research, we specifically focused on assessing gender bias in the language models using a downstream task. This choice was primarily influenced by the type of language model we employed, which is an encoder-only model commonly utilized in downstream tasks such as classification, named entity recognition (NER), and question answering (QA). For our evaluation, we utilized the sentiment analysis task. The subsequent section provides a detailed description of our methodology, outlining the steps involved.

\noindent

\subsection{Evaluating Gender Bias using Sentiment Analysis Task}

\begin{figure}
    \centering
    \scalebox{1.5}{
    \includegraphics{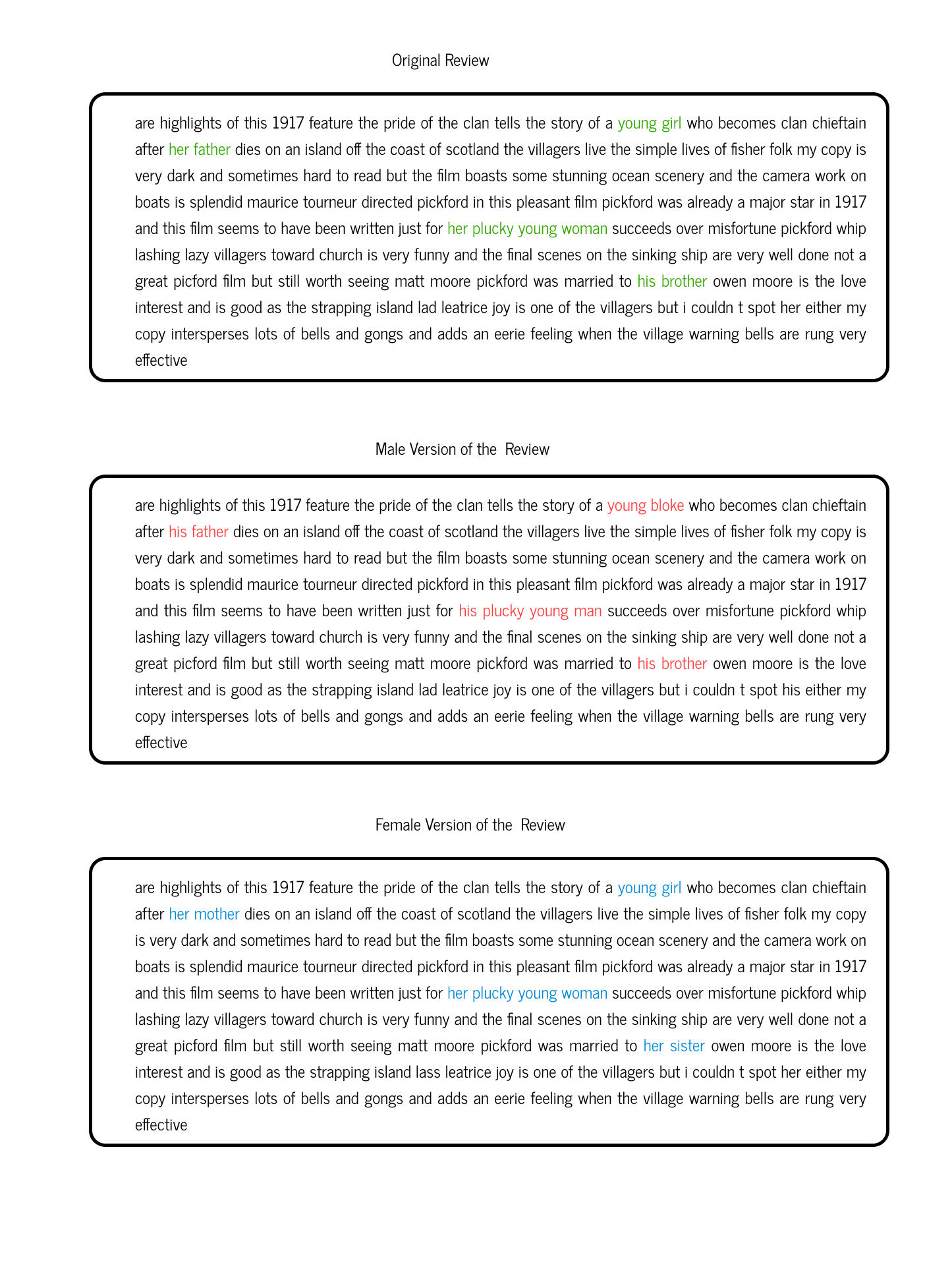}}
    \caption{Sample of a movie review along with two gender versions.}
    \label{fig:bias_sample}
\end{figure}

\noindent In order to analyze and compare the bias in the predictions of our sentiment analysis models, we adopt a methodology inspired by previous work on gender bias assessment \cite{jentzsch-turan-2022-gender}. We create two versions of each movie review from the IMDB dataset, representing the male and female genders, by substituting different keywords, names, and pronouns. The choice of gender-specific terms follows the same set used in the referenced study \cite{jentzsch-turan-2022-gender}. Figure \ref{fig:bias_sample} provides an example of a movie review with both male and female versions.

To quantify the bias in a sample's prediction, we calculate the difference in sentiment polarity between the male and female versions. This difference is computed based on the model's predicted sentiment scores for each version, as shown in Equation \ref{eq:bias_sample}. A higher bias value indicates that the model rated the sample as more positive for the male version compared to the female version.

\begin{eqnarray}
   \label{eq:bias_sample}
    Bias_i &=&sent(i_{male}) - sent(i_{female})  \\
    \label{eq:bias}
      Bias_{Model} &=& \sum_{i=1}^{N} \frac{ \triangle sent}{N} 
       \\
  \label{eq:abs_bias}
    Absolute Bias_{Model} &=& \sum_{i=1}^{N} \frac{ | \triangle sent |}{N}
\end{eqnarray}

To evaluate the overall bias of the models, we calculate the mean difference (as defined in Equation \ref{eq:bias}) and the mean absolute difference (as defined in Equation \ref{eq:abs_bias}) across all the samples in the validation set for each cross-validation fold of the IMDB data split. These measures provide an indication of the average bias present in the model's predictions.

To compare the bias between the two models, we perform a paired t-test. This statistical test allows us to assess whether there is a significant difference in the bias exhibited by the two models. It is important to note that our study's main focus is not to determine whether the models are biased or not, but rather to investigate if one model demonstrates significantly more bias than the other. By employing this methodology, we aim to provide a quantitative analysis of bias in the sentiment analysis models and explore any significant differences between them in terms of bias.

\chapter{RESULTS}
\label{chapter:results}

\section{Qualitative Comparison of Text Corpora}
\noindent We conducted a comprehensive analysis of the qualitative statistics for both the CNN/DailyMail and ChatGPT generated news articles, and the results are summarized in Table \ref{tab:stats}. In terms of word distribution, we ensured that the generated articles by ChatGPT had a similar number of words as the original CNN/DailyMail articles through the prompts. As a result, the word counts for both corpora are close to each other. However, we observed a significant difference in vocabulary size between the two. The vocabulary of the CNN/DailyMail articles was approximately 76\% larger compared to the ChatGPT articles. This indicates that the CNN/DailyMail corpus exhibited a wider range of unique words and phrases based on the statistics computed. 

Additionally, we found notable differences in the structural characteristics of the articles. The ChatGPT articles had a lower average number of sentences per article, around 16, compared to the CNN/DailyMail articles, which had an average of 20 sentences. Consequently, the sentences in the ChatGPT articles tended to be longer. Furthermore, we analyzed the presence of named entities in the articles. The mean number of named entities per article was nearly double in the CNN/DailyMail articles, with an average of around 41, compared to the ChatGPT articles, which had an average of 22 named entities. This indicates that the CNN/DailyMail articles contained a higher density of named entities.

Overall, these qualitative statistics provide insights into the differences between the CNN/DailyMail and ChatGPT articles, including vocabulary size, sentence structure, and the presence of named entities.

\begin{table}[]
    
    \centering
    \scalebox{0.9}{
    \begin{tabular}{lrr}
         & \textbf{CNN/DailyMail} & \textbf{ChatGPT}\\
         \hline
         \textbf{Number of Words} & 8,966,581 & 8,798,474\\
         \textbf{Mean words per article} & 358.66 & 351.93 \\
         \textbf{Vocabulary} & 159,105 & 90,145\\
         \textbf{Mean Vocabulary per article} & 192.47 & 179.01 \\
         \textbf{Mean Sentences per article} & 20.14 & 16.01 \\
         \textbf{Mean Words per sentence} & 19.29 & 22.40\\
         \textbf{Mean Named Entities per article} & 40.86 & 22.40 \\

         \hline
    \end{tabular}}
    \caption{Comparison of the statistics of CNN/DailyMail and ChatGPT articles.}
    \label{tab:stats}
\end{table}
\subsection{Sentiment Polarity}
\begin{figure}
    \centering
    \scalebox{0.4}{
    \includegraphics{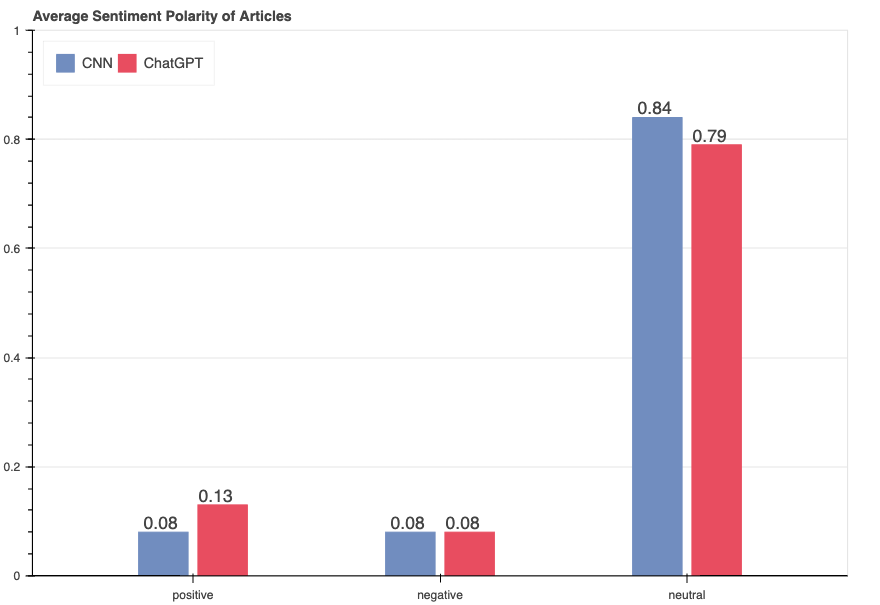}}
    \caption{Mean sentiment polarity of the articles using VADER Sentiment Analyzer.}
    \label{fig:senti}
\end{figure}

\noindent We conducted an analysis of the distribution of sentiment polarity in the articles from each source. Figure \ref{fig:senti} illustrates a comparison of the mean sentiment intensity between the CNN/DailyMail and ChatGPT articles.

In terms of sentiment intensity, we observe that the CNN/DailyMail articles have a higher neutral intensity compared to the articles written by ChatGPT. This suggests that the CNN/DailyMail articles exhibit a more balanced or neutral sentiment overall. On the other hand, the ChatGPT articles show a higher positive intensity compared to the CNN/DailyMail articles. This indicates that the ChatGPT-generated articles tend to have a more positive sentiment. Interestingly, the negative intensity is similar in both sets of articles, suggesting that the level of negativity is comparable between the CNN/DailyMail and ChatGPT articles.


\subsection{Distribution of Named Entities and POS tags}
\begin{figure}
    \centering
    \scalebox{0.4}{
    \includegraphics{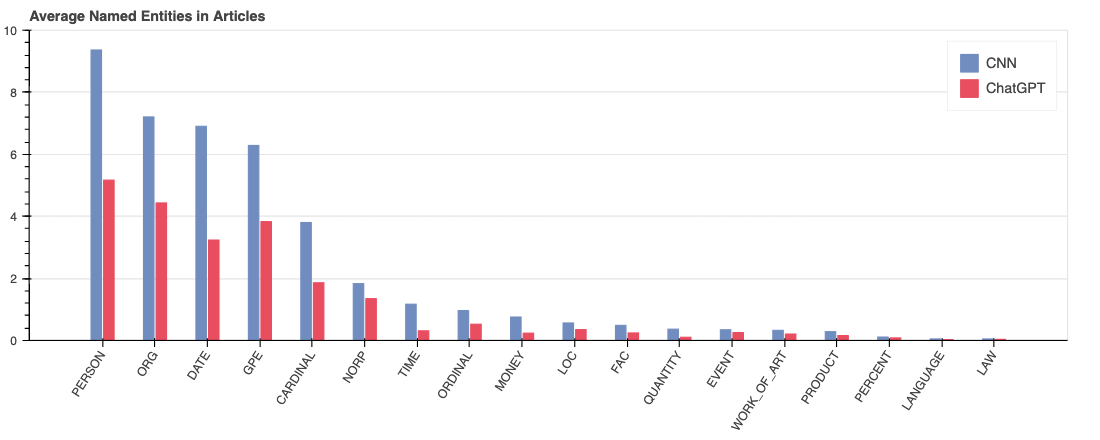}}
    \caption{Distribution of Named Entities in CNN/DailyMail and ChatGPT articles.}
    \label{fig:ent}
\end{figure}

\begin{figure}
    \centering
    \scalebox{0.4}{
    \includegraphics{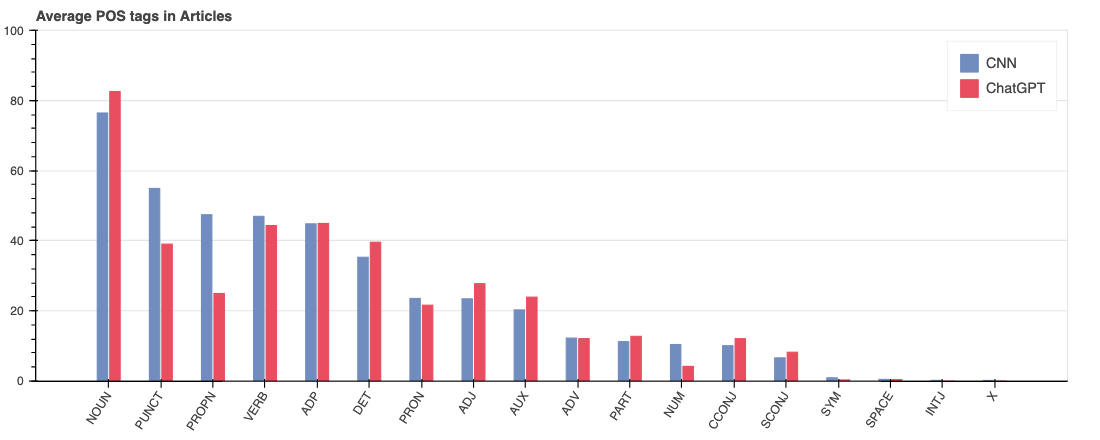}}
    \caption{Distribution of POS tags in CNN/DailyMail and ChatGPT articles.}
    \label{fig:pos}
\end{figure}

\noindent We conducted an analysis of the distribution of Named Entities and POS tags in the articles from each source. Figure \ref{fig:ent} displays the distribution of Named Entities in CNN/DailyMail articles, while Figure \ref{fig:pos} shows the distribution of POS tags in ChatGPT articles.

The analysis revealed interesting findings regarding the occurrence of Named Entities in the articles. Specifically, we observed that the number of named entities in ChatGPT articles was approximately half compared to the CNN/DailyMail articles. 

On the other hand, the distribution of POS tags, which provide information about the grammatical roles of words, showed a similar pattern between the two sources. This indicates that both CNN/DailyMail and ChatGPT articles exhibit comparable syntactic structures and grammatical usage. 

\subsection{Readability Metrics}
\noindent We measured two standard readability scores, Flesch-Reading Ease and Flesch-Kincaid Grade Level, on all the articles in the corpora. Table \ref{tab:read} provides the mean readability scores for each corpus.

\begin{figure}
    \centering
    \scalebox{0.4}{
    \includegraphics{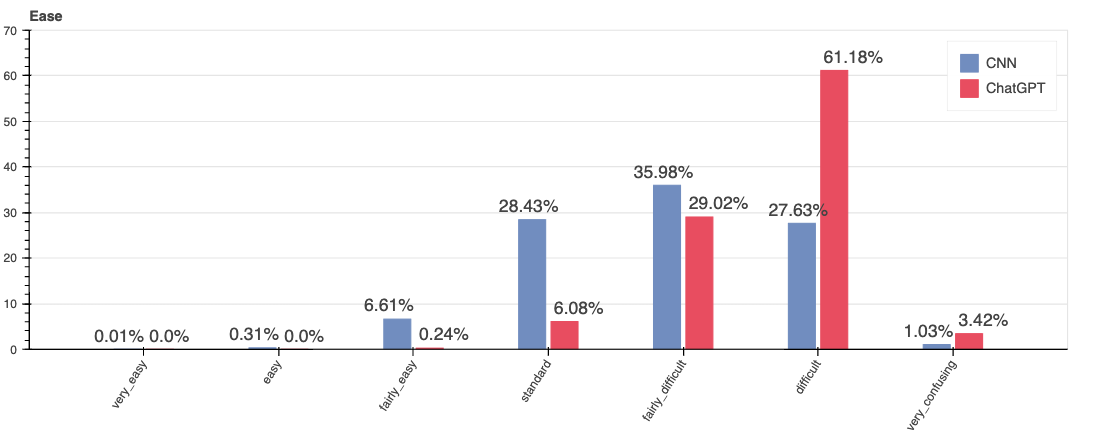}}
    \caption{Distribution of readability scores of CNN/DailyMail and ChatGPT articles using the Flesch Reading Ease.}
    \label{fig:ease}
\end{figure}

Flesch-Reading Ease scores are designed to indicate the difficulty level of a passage in English. Higher scores indicate that the text is easier to read, while lower scores indicate texts that are more challenging to comprehend. We observed that the articles written by ChatGPT have a Flesch-Reading Ease score of 46.49, compared to 55.52 for CNN/DailyMail. This suggests that the ChatGPT articles are more difficult to understand and require a higher level of education. To further illustrate the distribution of readability, we plotted the articles on a scale ranging from very easy to very confusing in Figure \ref{fig:ease}, based on the interpretation of Flesch-Reading Ease scores. We found that the ChatGPT articles were more skewed towards the difficult end, with over 93\% falling into that category, compared to about 64\% for CNN/DailyMail. 

Similarly, for the Flesch-Kincaid Grade Level, we observed that the articles written by ChatGPT were at a higher grade level by 2 points compared to CNN/DailyMail articles.

These findings suggest that the articles generated by ChatGPT tend to be more challenging in terms of readability, requiring a higher level of education to comprehend.
\begin{table}[]
    
    \centering
    \scalebox{0.9}{
    \begin{tabular}{lrr}
         & \textbf{CNN/DailyMail} & \textbf{ChatGPT}\\
         \hline
         \textbf{Mean Flesch-Reading Ease} & 55.52 & 46.49 \\
         \textbf{Mean Flesch–Kincaid Grade Level} & 10.11 & 12.11 \\

         \hline
    \end{tabular}}
    \caption{Readability scores of CNN/DailyMail and ChatGPT articles.}
    \label{tab:read}
\end{table}

\section{Pre-training}
\begin{figure}
    \centering
    \scalebox{0.15}{
    \includegraphics{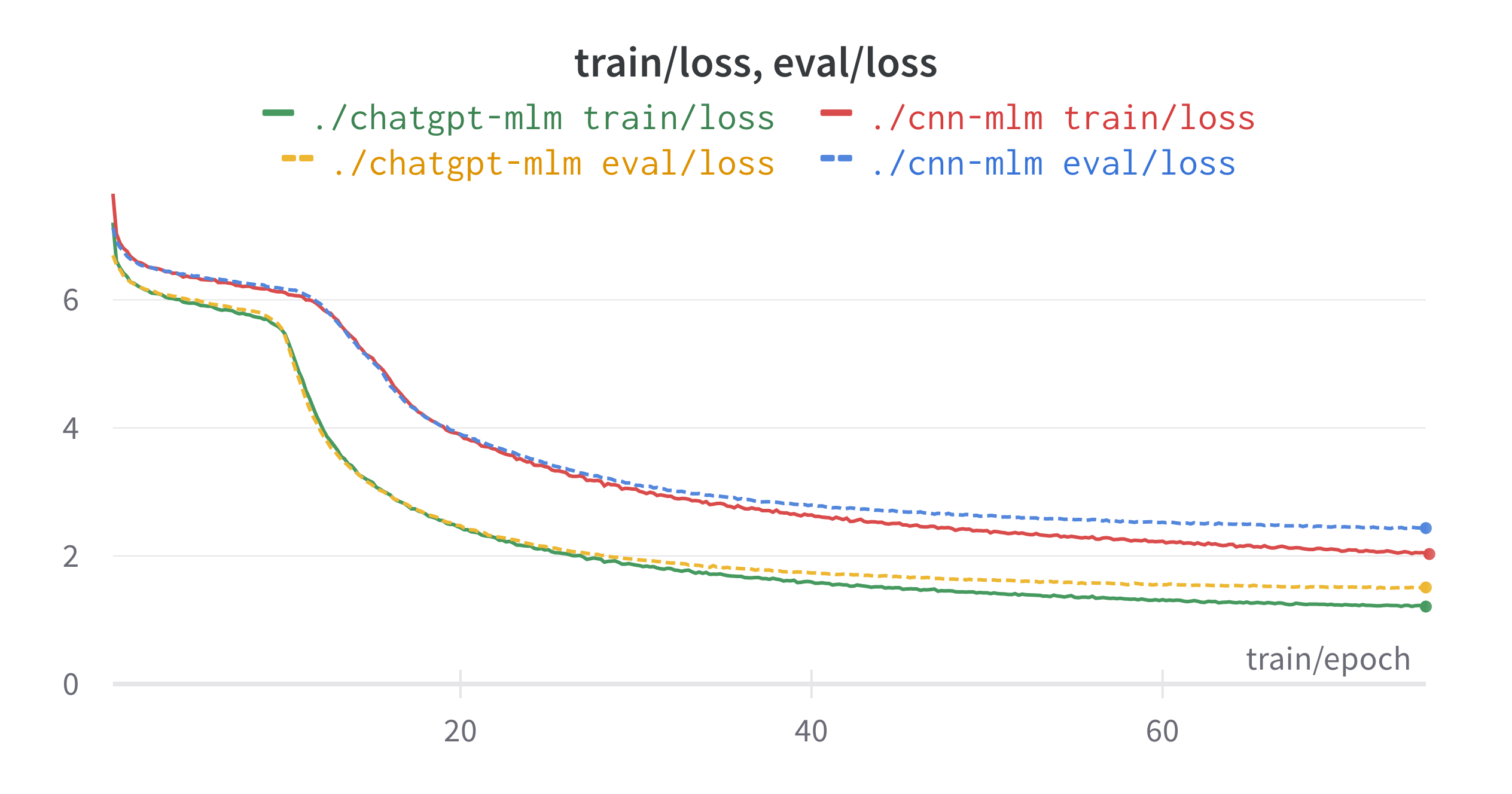}}
    \caption[Loss curves during RoBERTa pre-training on CNN/DailyMail and ChatGPT dataset.]{Loss curves during RoBERTa pre-training on CNN/DailyMail (red and blue) and ChatGPT dataset (green and yellow).}
    \label{fig:pre}
\end{figure}

\noindent The perplexity after pre-training the RoBERTa model on the CNN/DailyMail dataset and ChatGPT dataset is reported in Table \ref{tab:perpex}, and Figure \ref{fig:pre} displays the loss curves during pre-training. While it is important to note that the absolute values of loss/perplexity cannot be directly compared due to the use of different evaluation data (CNN/DailyMail text or ChatGPT text), certain observations can still be made. It is evident that the overall loss after 75 epochs is significantly lower for the ChatGPT articles compared to the CNN/DailyMail articles. Additionally, the loss curve for ChatGPT article pre-training consistently remains lower than that of the CNN/DailyMail articles. 
\begin{table}[]
    \centering
    \begin{tabular}{cr}
         \textbf{Model} & \textbf{Perplexity}  \\
         \hline
         \textbf{RoBERTa (CNN/DailyMail)} & 11.41 \\
         \textbf{RoBERTa (ChatGPT)} & 4.46 \\
         \hline
    \end{tabular}
    \caption[Perplexity of pre-trained RoBERTa models.]{Perplexity of pre-trained RoBERTa models on the evaluation dataset. (Note: the datasets are different) }
    \label{tab:perpex}
\end{table}
\section{Downstream Evaluation}
\subsection{Sequence Classification}
\noindent The results for the five iterations of two-fold cross-validation for each of the fine-tuned RoBERTa models, pre-trained on CNN/DailyMail and ChatGPT articles, are reported in Table \ref{tab:imdb-results}. The metric used for comparing the performance of the models on the IMDB classification task is accuracy. When comparing the average performance across all runs, we find that the RoBERTa model pre-trained on ChatGPT articles achieved a higher average accuracy of 0.858, compared to 0.851 for the RoBERTa model pre-trained on CNN/DailyMail articles. To establish a benchmark, we also fine-tuned the original RoBERTa model and obtained an average accuracy of 0.908 for the first iteration of splits.

Finally, to address our research sub-question 1(a), which is whether the RoBERTa (ChatGPT) model is significantly inferior in terms of accuracy on the sentiment classification task, we conducted a paired t-test to compare the two models. The computed t-statistics was -11.108, resulting in a p-value of 0.0001. Based on these statistics, we reject the null hypothesis that both models perform equally well on this task, as the p-value $(p < 0.0001)$ is smaller than the significance level $\alpha$ (0.05). Surprisingly, the RoBERTa (ChatGPT) model performed significantly better than the RoBERTa (CNN/DailyMail) model on the sentiment classification task.

\begin{table}[]
    \centering
    \begin{tabular}{|c|c|cc|}
    \hline
   \textbf{Iteration} & \textbf{Split} &   \textbf{RoBERTa (CNN/DailyMail)}   &  \textbf{RoBERTa (ChatGPT)} \\
   \hline
       \multirow{2}{*}{\#1}&Split 1 & 0.849 & \textbf{0.857}\\
    &Split 2 & 0.851& \textbf{0.858}\\ 
    \hline
    
       \multirow{2}{*}{\#2}&Split 1 & 0.851& \textbf{0.859}\\
    &Split 2 & 0.848& \textbf{0.855}\\ 
    \hline
    \multirow{2}{*}{\#3}&Split 1 & 0.852& \textbf{0.858}\\
    &Split 2 & 0.851&\textbf{0.858} \\ 
    \hline
    \multirow{2}{*}{\#4}&Split 1 & 0.851& \textbf{0.856}\\
    &Split 2 & 0.855& \textbf{0.860} \\ 
    \hline
    \multirow{2}{*}{\#5}&Split 1 & 0.854& \textbf{0.861}\\
    &Split 2 &0.849& \textbf{0.857} \\ 
    \hline
    \end{tabular}
    \caption{Results of the five iterations of cross validation on the IMDB classification challenge.}
    \label{tab:imdb-results}
\end{table}

\subsection{Named Entity Recognition}
\begin{table}[]
    \centering
    \begin{tabular}{|c|c|cc|}
    \hline
   \textbf{Iteration} & \textbf{Split} &   \textbf{RoBERTa (CNN/DailyMail)}   &  \textbf{RoBERTa (ChatGPT)} \\
   \hline
       \multirow{2}{*}{\#1}&Split 1 & \textbf{0.444} & 0.430\\
    &Split 2 & 0.448& \textbf{0.457}\\ 
    \hline
    
       \multirow{2}{*}{\#2}&Split 1 & \textbf{0.444}& 0.439\\
    &Split 2 & 0.440& \textbf{0.442}\\ 
    \hline
    \multirow{2}{*}{\#3}&Split 1 & 0.439& \textbf{0.448}\\
    &Split 2 & \textbf{0.431}&0.424 \\ 
    \hline
    \multirow{2}{*}{\#4}&Split 1 & 0.426& \textbf{0.429}\\
    &Split 2 & \textbf{0.458}& 0.444 \\ 
    \hline
    \multirow{2}{*}{\#5}&Split 1 & \textbf{0.456}& 0.439\\
    &Split 2 &0.439& \textbf{0.444} \\ 
    \hline
    \end{tabular}
    \caption{Results of the five iterations of cross validation on the WNUT-17 Named Entity Recognition Task.}
    \label{tab:ner-results}
\end{table}

\noindent The results for the five iterations of cross-validation for each of the fine-tuned RoBERTa models using pre-trained CNN/DailyMail and ChatGPT articles are presented in Table \ref{tab:ner-results}. The metric used to evaluate the model performances on the WNUT-17 NER task is F1-score. When considering the average performance across all iterations, we observe that the RoBERTa model pre-trained on ChatGPT achieves a similar F1-score of 0.440, compared to 0.442 for the RoBERTa model pre-trained on CNN/DailyMail. To establish a benchmark, we also fine-tune the original RoBERTa model, obtaining an average F1-score of 0.491 for the first iteration of splits.

Moving on to address our research sub-question 1(b), which examines whether the RoBERTa (ChatGPT) model is significantly inferior in terms of F1-score on the Named Entity Recognition task, we conducted a paired t-test to compare the two models. The computed t-statistics yielded a value of 1.101, resulting in a corresponding p-value of 0.321.

Based on these statistical findings, we fail to reject the null hypothesis and conclude that there is no significant difference in performance between the two models on the Named Entity Recognition task, as the p-value $(p > 0.321)$ is greater than the predefined significance level $\alpha$ (0.05). Thus, we find no evidence to suggest that the RoBERTa (ChatGPT) model is significantly inferior in terms of F1-score on the Named Entity Recognition task.

\subsection{Question Answering}

\begin{table}[]
    \centering
    \begin{tabular}{|c|c|cc|}
    \hline
   \textbf{Iteration} & \textbf{Split} &   \textbf{RoBERTa (CNN/DailyMail)}   &  \textbf{RoBERTa (ChatGPT)} \\
   \hline
       \multirow{2}{*}{\#1}&Split 1 & 63.906 & \textbf{64.880}\\
    &Split 2 & 63.896& \textbf{64.981}\\ 
    \hline
    
       \multirow{2}{*}{\#2}&Split 1 & 64.086& \textbf{64.636}\\
    &Split 2 & 63.506& \textbf{65.078} \\ 
    \hline
    \multirow{2}{*}{\#3}&Split 1 & 64.425& \textbf{64.921}\\
    &Split 2 & 64.436& \textbf{64.987} \\ 
    \hline
    \multirow{2}{*}{\#4}&Split 1 & 64.266& \textbf{65.398}\\
    &Split 2 & 64.349& \textbf{64.972} \\ 
    \hline
    \multirow{2}{*}{\#5}&Split 1 & 64.971& \textbf{65.050}\\
    &Split 2 & 63.713& \textbf{64.779} \\ 
    \hline
    \end{tabular}
    \caption{Results of the five iterations of cross validation on the on SQuAD QA Challenge.}
    \label{tab:squad-results}
\end{table}

\noindent The results for the five iterations of cross-validation for each of the fine-tuned RoBERTa models using pre-trained CNN/DailyMail and ChatGPT articles are presented in Table \ref{tab:squad-results}. The metric used to evaluate the model performances on the SQuAD Question Answering task is F1-score. When considering the average performance across all iterations, we observe that the RoBERTa model pre-trained on ChatGPT achieves a similar F1-score of 64.968, compared to 64.156 for the RoBERTa model pre-trained on CNN/DailyMail. To establish a benchmark, we also fine-tune the original RoBERTa model, obtaining an average F1-score of 83.80 for the first iteration of splits.

Moving on to address our research sub-question 1(c), which examines whether the RoBERTa (ChatGPT) model is significantly inferior in terms of F1-score on the QA task, we conducted a paired t-test to compare the two models. The computed t-statistics yielded a value of -2.035, resulting in a corresponding p-value of 0.097.

Based on these statistical findings, we fail to reject the null hypothesis and conclude that there is no significant difference in performance between the two models on the QA task, as the p-value $(p > 0.097)$ is greater than the predefined significance level $\alpha$ (0.05). Thus, we find no evidence to suggest that the RoBERTa (ChatGPT) model is significantly inferior in terms of F1-score on the QA task.

\section{Evaluation of Gender Biases}

\noindent Based on our definitions of overall bias in a model, we computed the results for the five iterations of cross-validation models, which are presented in Table \ref{tab:mean_bias}. The bias values range from -1 to +1, where a value of -1 indicates that the model assigns more positive polarity to female versions of the samples, and a value of +1 indicates that the model assigns more positive polarity to male versions of the samples. In all cases, we observed that the bias values were positive, indicating a tendency to assign more positive polarity to male versions.

Overall, in half of the cases, the RoBERTa model pre-trained on ChatGPT exhibited a lower bias value compared to the RoBERTa model pre-trained on CNN/DailyMail. However, it is worth noting that the mean bias value for RoBERTa (CNN/DailyMail) was 0.0056, slightly higher than the mean bias value for RoBERTa (ChatGPT), which was 0.0053.

Finally, to address our research question 2, which examines whether the RoBERTa (ChatGPT) model is significantly more biased towards gender demographic, we conducted a paired t-test to compare the two models. The computed t-statistics yielded a value of 1.712, resulting in a corresponding p-value of 0.147. Based on these statistical findings, we fail to reject the null hypothesis and conclude that there is no significant difference in bias performance between the two models, as the p-value $(p > 0.147)$ is greater than the predefined significance level $\alpha$ (0.05). Thus, we find no evidence to suggest that the RoBERTa (ChatGPT) model is not significantly more biased for the gender demographic.

\begin{table}[]
    \centering
    \begin{tabular}{|c|c|cc|}
    \hline
   \textbf{Iteration} & \textbf{Split} &   \textbf{RoBERTa (CNN/DailyMail)}   &  \textbf{RoBERTa (ChatGPT)} \\
   \hline
       \multirow{2}{*}{\#1}&Split 1 & 0.0060& \textbf{0.0042}\\
    &Split 2 & \textbf{0.0052}&0.0059 \\ 
    \hline
    
       \multirow{2}{*}{\#2}&Split 1 & 0.0065&\textbf{0.0047}\\
    &Split 2 & 0.0048& \textbf{0.0024} \\ 
    \hline
    \multirow{2}{*}{\#3}&Split 1 & \textbf{0.0044}&0.0048\\
    &Split 2 & \textbf{0.0056}&0.0059 \\ 
    \hline
    \multirow{2}{*}{\#4}&Split 1 & 0.0074&\textbf{0.0067}\\
    &Split 2 & 0.0041& \textbf{0.0025} \\ 
    \hline
    \multirow{2}{*}{\#5}&Split 1 & \textbf{0.0085}&0.0109\\
    &Split 2 & \textbf{0.0040} &0.0045 \\ 
    \hline
    \end{tabular}
    \caption{Mean difference between male and female polarity sentiments.}
    \label{tab:mean_bias}
\end{table}

In addition to above results, we also measured the absolute difference in polarities, which is reported in Table \ref{tab:abs_bias}. In terms of the absolute score, the RoBERTa model pre-trained on ChatGPT had a lower score 80\% of the time compared to the RoBERTa model pre-trained on CNN/DailyMail.


\begin{table}[]
    \centering
    \begin{tabular}{|c|c|cc|}
    \hline
   \textbf{Iteration} & \textbf{Split} &   \textbf{RoBERTa (CNN/DailyMail)}   &  \textbf{RoBERTa (ChatGPT)} \\
   \hline
       \multirow{2}{*}{\#1}&Split 1 & 0.0124& \textbf{0.0104}\\
    &Split 2 & \textbf{0.0100}&0.0117 \\ 
    \hline
    
       \multirow{2}{*}{\#2}&Split 1 & 0.0111& \textbf{0.0109}\\
    &Split 2 & 0.0113& \textbf{0.0094} \\ 
    \hline
    \multirow{2}{*}{\#3}&Split 1 & 0.0126& \textbf{0.0118}\\
    &Split 2 & 0.0120& \textbf{0.0112} \\ 
    \hline
    \multirow{2}{*}{\#4}&Split 1 & 0.0131& \textbf{0.0125}\\
    &Split 2 & 0.0100& \textbf{0.0099} \\ 
    \hline
    \multirow{2}{*}{\#5}&Split 1 & \textbf{0.0132}&0.0159\\
    &Split 2 & 0.0109& \textbf{0.0106} \\ 
    \hline
    \end{tabular}
    \caption{Mean absolute difference between male and female polarity sentiments.}
    \label{tab:abs_bias}
\end{table}


\chapter{DISCUSSION AND LIMITATIONS}


\section{Discussion}
\noindent In the comparison of writing styles between CNN/DailyMail journalists and ChatGPT, significant differences were observed. The vocabulary diversity was notably higher for journalists compared to ChatGPT (the temperature of ChatGPT was not explored). This factor could potentially explain the lower perplexity and loss curves in the MLM task for the text generated by ChatGPT. Additionally, the articles generated using ChatGPT exhibited a more positive sentiment compared to the slightly higher neutral sentiment observed in journalist-written articles. This could be attributed to the positive human feedback received during the alignment of ChatGPT using reinforcement learning \cite{bai2022training}.

Regarding grammatical structure and complexity measured using readability metrics, the articles written by ChatGPT were found to be significantly more complex and required a higher grade level for understanding compared to journalist-written articles. This may provide a rationale for why RoBERTa trained on ChatGPT articles performed well in both pre-training and downstream tasks, as it effectively learned the intricacies of the language. Further details regarding the performance of downstream tasks and bias analysis will be discussed in the subsequent sections.

\begin{figure}
\centering
\scalebox{0.4}{
    \includegraphics{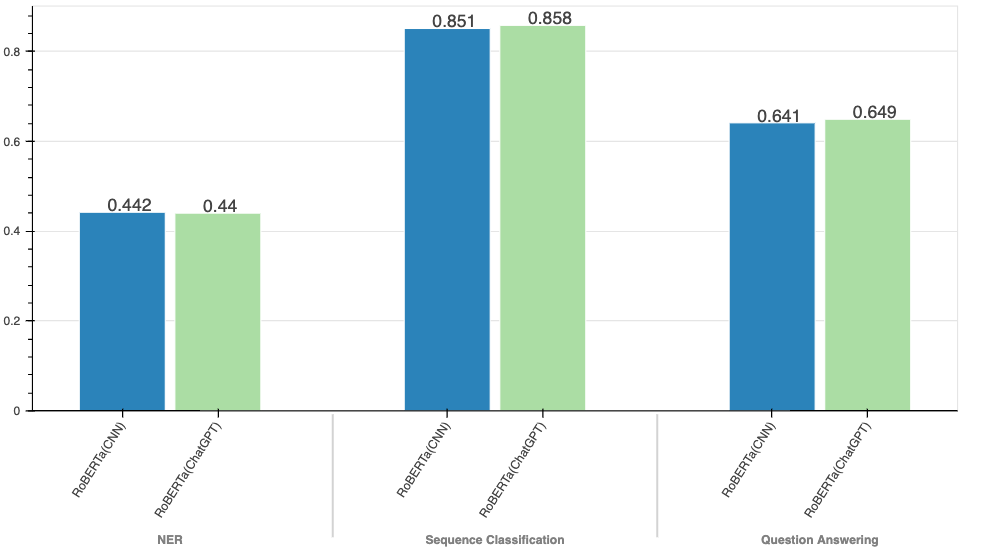}}
    \caption[Mean performance of models across different tasks.]{Mean performance of models across different tasks (accuracy for sequence classification task and F1-score in others).}
    \label{fig:perf}
\end{figure}

\section{Downstream Performance}
\noindent 
To compare the performance of the two RoBERTa models trained on ChatGPT and CNN/DailyMail articles, we selected three tasks: sequence classification, named entity recognition (NER), and question answering. Each task consisted of ten runs for each model, and a summary of the wins across the different tasks is presented in Table \ref{tab:summary}. Additionally, Figure \ref{fig:perf} displays the mean F1 score of all the runs for each task.

Surprisingly, the RoBERTa (ChatGPT) model achieved more wins and demonstrated better performance in two of the tasks: sequence classification and question answering. However, the performance improvement was statistically significant only for the sequence classification task, as confirmed by the t-test. In the NER task, both RoBERTa models achieved an equal number of wins, but the RoBERTa (CNN/DailyMail) model exhibited slightly better mean performance across all the runs, as depicted in Figure \ref{fig:perf}.


\begin{table}[]
    \centering
    \scalebox{0.8}{
    
    \begin{tabular}{lccc}
        \textbf{Task} & \textbf{RoBERTa(CNN) Wins} & \textbf{RoBERTa(ChatGPT) Wins} & \textbf{T-Test}   \\
        \hline
        Sequence Classification & 0 & 100\% & Significant \\
        NER Task & 50\% & 50\% & Insignificant \\
        Question Answering & 0\% & 100\% & Insignificant \\
        \hline
    \end{tabular}}
    \caption{Summary of the performance of different models on different tasks.}
    \label{tab:summary}
\end{table}

\section{Bias Performance}
\noindent To examine the inherent bias in our trained models, we measured the sentiment polarity difference between male and female versions of the instances. The overall bias of the models was represented using the mean difference and mean absolute difference. The summary of the bias results for our models is presented in Table \ref{tab:summary}.

In terms of the mean difference, both models performed similarly across all ten cases, indicating no significant bias favoring either male or female versions. However, when considering the absolute mean difference, the RoBERTa (ChatGPT) model exhibited lower bias in eight out of ten cases compared to the RoBERTa (CNN/DailyMail) model. It is important to note that in both cases, the difference in bias was statistically insignificant.

\begin{table}[]
    \centering
    \scalebox{0.8}{
    
    \begin{tabular}{lccc}
        \textbf{Bias Metric} & \textbf{RoBERTa(CNN) Wins} & \textbf{RoBERTa(ChatGPT) Wins} & \textbf{T-Test}   \\
        \hline
        Mean Bias & 50\% & 50\% & Insignificant \\
        Absolute Mean Bias & 20\% & 80\% & Insignificant \\
        
        \hline
    \end{tabular}}
    \caption{Summary of the Bias performance of different models.}
    \label{tab:summary-bias}
\end{table}

\section{Limitations}
\label{chapter:limitations}
\noindent While our research provides valuable insights, it is important to acknowledge the limitations inherent in our study:
\begin{enumerate}
    \item \textbf{Limited to a specific data source}: Our research primarily centres around a specific data source, namely the news articles from the CNN/DailyMail dataset. It is important to note that language models are typically trained on much larger corpora that encompass a wide range of data sources, including books, Wikipedia, and various social media platforms such as Facebook, Twitter, and Reddit. The use of a specific dataset in our study may not fully capture the diversity of domains and topics present in the training and generation of language models. Consequently, there is a potential risk that the limited scope of our study may restrict the generalizability of our findings to other datasets or sources. This limitation could potentially impact the broader applicability of our results and the extent to which our conclusions can be applied to different contexts.
    \item \textbf{Text Decoding}: The creativity and diversity exhibited by a language model during the text generation process can be influenced by various factors, including the decoding temperature and strategies employed \cite{holtzman2019curious}. In our study, we utilized the default temperature parameter and decoding strategy, which may have resulted in lower diversity in terms of vocabulary and Named Entity Recognition (NER) in the generated articles. We recognize that the choice of decoding strategy can significantly impact the output of the language model, and it is an important area for further investigation. Future studies could explore the impact of different decoding strategies and temperature settings and pre-train models on more creative and diverse generated text, providing a more comprehensive understanding of their capabilities and limitations.
    \item \textbf{Size of pre-training dataset}: The size of the pre-training dataset plays a crucial role in determining the performance of language models. In our study, we utilized a relatively small pre-training dataset consisting of 10 million tokens, whereas contemporary models are often trained on billions of tokens. While some studies have demonstrated that smaller language models can learn effectively from datasets in the order of millions of tokens, it would be valuable to examine whether our findings hold true for larger language models as well. Conducting experiments with a larger pre-training dataset would provide further insights into the performance and generalizability of language models, allowing for a more comprehensive understanding of their capabilities and limitations.

    \item \textbf{Sensitivity of prompting and steering}: The process of text generation by language models is known to be influenced by the prompts provided. The selection of prompts plays a crucial role in shaping the output and has the potential to introduce biases or limitations. In our study, we did not specifically investigate the effects of steering the outputs of ChatGPT using different prompts. However, the users of language models have the ability to steer the results by providing specific prompts. This raises the possibility that individuals could intentionally or unintentionally generate outputs that are unfair, biased, or unrepresentative. Exploring the impact of different prompting strategies on bias metrics is an area that could be further explored in future research.
    \item \textbf{Language restriction to English}: Our study is limited to the analysis of the English language only. This restriction prevents us from investigating language-specific nuances, biases, and behaviors that may vary in other languages. By expanding the scope of our analysis to include multiple languages, we would gain a more comprehensive understanding of how language models perform and exhibit biases in different linguistic contexts. Such a multilingual approach would enhance the generalisability and applicability of our findings to a wider range of languages and cultures.
    \item \textbf{Exclusive use of ChatGPT}: 
Our research exclusively focuses on the ChatGPT-03 language model, which incorporates protective measures to mitigate the generation of biased and harmful content. We do not investigate the impact of uncensored language models that operate without relying on human feedback for safety, such as the Wizard-Vicuna-30B-Uncensored-GGML model\footnote{https://huggingface.co/TheBloke/Wizard-Vicuna-30B-Uncensored-GGML}. This choice restricts our ability to fully comprehend how different language models may influence our results and findings, and it may limit the broader analysis of language generation models as a whole. Future research could explore the implications and variations that arise from utilizing diverse language models with varying levels of content moderation.
    \item \textbf{Bias Metrics}: While our research focuses on gender bias through sentiment analysis, it does not encompass the examination of other forms of bias, such as racial or cultural biases. Additionally, there exist diverse methodologies and definitions for evaluating biases in language models. Exploring these various forms of biases and employing different evaluation techniques could be a valuable avenue for future research, especially as language models continue to evolve and advance. By expanding the scope of bias analysis, we can gain a more comprehensive understanding of the potential biases present in language models and develop strategies for addressing them effectively.
    
\end{enumerate}
\noindent Despite the inherent limitations of our research, we believe that our study provides valuable insights to the research community. While we acknowledge the boundaries and constraints within our research design, we maintain that our findings and conclusions hold significance and contribute to the understanding of language models. The insights gained from our study have the potential to guide future research in the development of language models, addressing potential biases, ensuring robust and reliable performance, and addressing concerns related to privacy, security, and fairness. Our research aims to contribute to the broader discussion surrounding language models and foster advancements in the field to promote the responsible and beneficial use of these models.

\chapter{ETHICAL AND ENVIRONMENTAL IMPACTS}
\section{Ethical Implications}
The impact of AI on our society is widely recognized as significant, and therefore, research and findings related to our understanding of these systems can have both positive and negative implications. In conducting our study, we acknowledge the ethical implications that arise from our research and findings. We recognize the importance of considering the following aspects with care:
\begin{enumerate}
    \item \textbf{Generalizability}: Based on our findings, we report that pre-training the language model on the dataset generated using ChatGPT did not significantly impact its performance on downstream tasks or its bias towards the gender demographic. However, we acknowledge the limitations in the generalizability of our findings, and further studies are necessary to thoroughly assess the extent of this impact.
    \item \textbf{Potential for bias/discrimination}: 
During the generation of articles in our study, we did not manipulate or control the tone or emotion of the news articles. However, we recognize the potential negative consequences that can arise if steering techniques are employed to manipulate the output of language models. It is important to acknowledge and address these ethical concerns to ensure responsible use of such techniques.
    
\end{enumerate}

\section{Environmental Impacts}
\noindent 

Deep learning models, particularly large transformer models like GPT, require substantial computational resources for training. This high computational demand leads to significant energy consumption, contributing to carbon emissions. The environmental impact of training machine learning models has been highlighted in various studies \cite{strubell-etal-2019-energy, bender2021dangers, report-carbon}, emphasizing the importance of assessing and understanding these impacts.

In line with these concerns, we acknowledge the significance of studying the carbon footprint associated with training language models. In our research, we have taken into account the carbon emissions generated by our experiments. To quantify and report the carbon footprints, we have utilized an open tool \footnote{https://mlco2.github.io/impact/} developed by Lacoste et al. \cite{lacoste2019quantifying}. The tool allows us to assess the environmental impact and estimate the carbon emissions associated with our experiments. By considering the carbon footprints of our research, we aim to contribute to the ongoing dialogue on the environmental sustainability of machine learning practices and encourage further investigation into reducing the carbon footprint of training large language models.

\begin{table}[]
    \centering
    \begin{tabular}{crr}
         \textbf{Task} & \textbf{Dataset} & \textbf{Carbon Emission} $(kg CO_2)$ \\
         \hline
         Pre-training & CNN-DailyMail & 1.78 \\
         Pre-training & ChatGPT & 1.55 \\
         Fine-tuning & IMDB & 0.23\\
         Fine-tuning & WNUT-17 & 0.02\\
         Fine-tuning & SQuAD & 0.57\\
         \hline
        Total & & 4.10\\
        \hline
    \end{tabular}
    \caption{Quantity of carbon emission associated with our experiments. }
    \label{tab:carbon_ems}
\end{table}
    
In our study, we conducted our experiments using the Google Cloud Platform (GCP) located in the Europe-West1 region, which has a carbon efficiency of 0.27 kgCO$_2$eq/kWh \cite{lacoste2019quantifying}. The carbon emissions associated with our experiments are reported in Table \ref{tab:carbon_ems}. 

The pre-training step of our language models contributed the most significant amount of carbon emissions. Pre-training on the CNN/DailyMail articles resulted in approximately 1.78 kg of CO$_2$ emissions, which is equivalent to 0.89 kg of coal burned \cite{chen2005technical} or driving an average Internal Combustion Engine car for 7.19 km \cite{epa2022inventory}. Pre-training on the ChatGPT articles had the second-highest carbon emissions of about 1.55 kg of CO$_2$, which is equivalent to 0.78 kg of coal burned or driving an average Internal Combustion Engine car for 6.26 km. 

Comparatively, fine-tuning had lower carbon emissions, even when we performed testing 20 times for each of the tasks (sentiment analysis, named-entity recognition, question-answering). The carbon emissions for the question-answering, sentiment analysis, and named-entity recognition tasks were 0.57 kg CO$_2$, 0.23 kg CO$_2$, and 0.02 kg CO$_2$, respectively. 

Overall, the total carbon emissions from all of our experiments amounted to approximately 4.1 kg of CO$_2$. It is important to note that since the data center was located in Europe, our emissions were offset by the cloud provider, mitigating the environmental impact to some extent. These carbon emissions data provide insights into the environmental footprint associated with training language models and highlight the need for considering sustainability aspects in machine learning research and practices.

\chapter{CONCLUSION}
\label{chapter:conclusion}
In conclusion, our research aimed to investigate the impact of artificial text on the performance of the RoBERTa language model. To address this research problem, we formulated two research questions. The first question focused on determining whether the RoBERTa model, pre-trained using artificial text, exhibited inferior performance across three different tasks. Our experimental results revealed no evidence to suggest that the RoBERTa (ChatGPT) model had inferior performance in any of the tasks. In fact in the sentiment analysis task, the model pre-trained on the artificial text had a better performance which is quite surprising. This opens a lot of new questions such as studying why and where artificial text can be an advantage. 

The second research question aimed to examine whether the RoBERTa (ChatGPT) model exhibited more bias towards different genders compared to the RoBERTa (CNN/DailyMail) model. However, our findings showed no evidence to support the notion that the RoBERTa (ChatGPT) model had greater bias than the RoBERTa (CNN/DailyMail) model. But given the fact that we only evaluated the gender bias, we believe that it is equally important to do further studies for evaluating other types of biases such as race, occupation etc.

Overall, our research did not demonstrate the presence of a significant impact of using artificial text for pre-training the language model. However, it is important to note that further research is needed to explore this topic in greater depth and across a broader range of experiments and datasets to test the generalizability. It is also important to acknowledge that such large-scale experiments should be conducted responsibly considering the impacts on humans and the environment.

\bibliographystyle{fbe_tez_v11}
\bibliography{references}

\appendix

\chapter{Training Details}
\label{parameters}
\noindent This section presents a comprehensive overview of our model training process. The complete codebase associated with our research can be accessed and retrieved from the GitHub repository\footnote{Github repo; https://github.com/sarthusarth/lang\_mod\_chatgpt}. By providing access to our codebase, we aim to facilitate reproducibility and transparency in our research methodology. The GitHub repository contains the necessary scripts, configurations, and dependencies required to reproduce the model training process described in this study. In addition to providing access to our codebase, we also open-source our trained models for testing purposes.
\section{Pre-training Details} 
\noindent All the hyperparameters associated with the per-taining of the RoBERTa models on both the datasets (CNN/DailyMail and ChatGPT articles) are reported in the table \ref{tab:hyperparameters}.

\begin{table}[]
    \centering
    \begin{tabular}{lr}
    
       \textbf{Hyperparameter}  & \textbf{Value} \\
       \hline
        Learning Rate & 5e-5 \\
        Batch Size & 8 \\
        Eval Batch Size & 8 \\
        Seed & 42 \\
        Optimizer &  Adam with betas(0.9,0.999) and epsilon(1e-08) \\ 
        LR Scheduler Type & Linear \\
        LR Scheduler Warmup Steps& 6 \\
        Number of Epochs & 75 \\
        \hline
    \end{tabular}
    \caption{Hyperparamters for Pre-Training Language Models.}
    \label{tab:hyperparameters}
\end{table}

\section{Fine-tuning Details on NER Task} 
\noindent All the hyperparameters associated with the fine-tuning both of the RoBERTa models on WNUT-17 NER dataset are reported in the table \ref{tab:hyperparameters-ner}.

\section{Fine-tuning Details on Classification Task} 
\noindent All the hyperparameters associated with the fine-tuning both of the RoBERTa models on IMDB dataset are reported in the table \ref{tab:hyperparameters-imdb}.

\section{Fine-tuning Details on Question-Answering Task} 
\noindent All the hyperparameters associated with the fine-tuning both of the RoBERTa models on SQuAD dataset are reported in the table \ref{tab:hyperparameters-squad}.

\begin{table}[]
    \centering
    \begin{tabular}{lr}
       \textbf{Hyperparameter}  & \textbf{Value} \\
       \hline
        Learning Rate & 5e-5\\
        Batch Size & 128 \\
        Eval Batch Size & 128 \\
        Optimizer &  Adam with betas(0.9,0.999) and epsilon(1e-08) \\ 
        LR Scheduler Type & Linear \\
        Number of Epochs & 3 \\
        \hline
    \end{tabular}
    \caption{Hyperparamters for WNUT-17 NER Task.}
    \label{tab:hyperparameters-ner}
\end{table}

\begin{table}[]
    \centering
    \begin{tabular}{lr}
       \textbf{Hyperparameter}  & \textbf{Value} \\
       \hline
        Learning Rate & 5e-5\\
        Batch Size & 128 \\
        Eval Batch Size & 128 \\
        Optimizer &  Adam with betas(0.9,0.999) and epsilon(1e-08) \\ 
        LR Scheduler Type & Linear \\
        Number of Epochs & 2 \\
        \hline
    \end{tabular}
    \caption{Hyperparamters for IMDB Sentiment Classification Task.}
    \label{tab:hyperparameters-imdb}
\end{table}

\begin{table}[]
    \centering
    \begin{tabular}{lr}
       \textbf{Hyperparameter}  & \textbf{Value} \\
       \hline
        Learning Rate & 3e-5\\
        Batch Size & 48 \\
        Eval Batch Size & 48 \\
        Optimizer &  Adam with betas(0.9,0.999) and epsilon(1e-08) \\ 
        LR Scheduler Type & Linear \\
        Number of Epochs & 2 \\
        Doc Stide Length & 128 \\
        Max Sequence Length & 384 \\
        \hline
    \end{tabular}
    \caption{Hyperparamters for SQuAD QA Task.}
    \label{tab:hyperparameters-squad}
\end{table}

\end{document}